\documentclass[sigconf]{acmart}

\usepackage{microtype}
\usepackage{graphicx}
\usepackage{subfigure}
\usepackage{hyperref}

\usepackage[capitalize,noabbrev]{cleveref}

\theoremstyle{plain}
\newtheorem{theorem}{Theorem}[section]

\theoremstyle{definition}

\theoremstyle{remark}

\usepackage{color-edits}
\addauthor{jiri}{red}
\addauthor{illia}{purple}

\newcommand{\debug}[1]{#1}		
\newcommand{\newmacro}[2]{\newcommand{#1}{\debug{#2}}}		

\newcommand{\RR}{ \mathbb{R} }

\newcommand{\spaceo}{\hspace{2 mm}}
\newcommand{\setsep}{ \spaceo | \spaceo}
\newcommand{\half}{\frac{1}{2}}

\newcommand{\Probu}[2]{\mathbb{P}_{#1}\left( #2 \right)}

\newcommand{\Abs}[1]{\left| #1 \right|}
\newcommand{\Set}[1]{\left\{ #1 \right\}}
\newcommand{\Brack}[1]{\left( #1 \right)}

\newcommand{\Expsubidx}[2]{ \mathbb{E}_{#1} #2}

\newcommand{\norm}[1]{\left\|#1\right\|}

\newlength{\dhatheight}

\newmacro{\ndata}{n}
\newmacro{\nsamples}{\ndata}
\newmacro{\ndims}{d}
\newmacro{\dims}{[\ndims]}
\newmacro{\nprot}{p}
\newmacro{\distance}{\Delta}

\newmacro{\prot}{\mathcal{P}}
\newmacro{\nonprot}{\mathcal{N}}
\newmacro{\target}{Y}

\newmacro{\subg}{S}
\newcommand{\subgs}[1][\prot]{\debug{\mathcal{S}_{#1}}}
\newmacro{\Xsubg}{\mathcal{X}_\subg}

\newmacro{\positive}{+}
\newmacro{\negative}{-}

\newmacro{\datadistr}{D}
\newmacro{\posdistr}{\mu}
\newmacro{\negdistr}{\nu}

\newmacro{\data}{\mathcal{D}}
\newmacro{\datapos}{\mathcal{D}_\positive}
\newmacro{\dataneg}{\mathcal{D}_\negative}

\newmacro{\error}{e}

\newmacro{\usevariable}{u}
\newcommand{\usevar}[1][j]{\usevariable_{#1}}

\newmacro{\xvariable}{x}

\newmacro{\minSize}{N_{\text{min}}}

\newmacro{\yhat}{\hat{y}_i}

\newmacro{\subgvariable}{s}

\newmacro{\accvariable}{a}

\newmacro{\accref}{\accvariable_\text{REF}}

\newmacro{\accrefout}{\accvariable^{\prime}_\text{REF}}

\newmacro{\DNF}{DNF}
\newmacro{\nterms}{t}

\newcommand{\MSD}[4][\distance]{\debug{\textrm{MSD}}_{#1}(#2,#3;#4)}
\newmacro{\MSDdiff}{\MSD[]\posdistr\negdistr\prot}

\newcommand{\MSDc}{\debug{\mathrm{MSD}}}

\newcommand{\istrue}[1]{\left\llbracket #1 \right\rrbracket}

\newcommand{\bigoh}[1]{\debug{\mathrm{O}}(#1)}
\newcommand{\bigomega}[1]{\debug{\mathrm{\Omega}}(#1)}

\newmacro{\cond}{\,\mid\,}

\newcommand{\mcX}{\mathcal{X}}
\newcommand{\mcF}{\mathcal{F}}

\newcommand{\mcL}{\mathcal{L}}
\newcommand{\mcB}{\mathcal{B}}

\usepackage{xcolor}

\author{Jiří Němeček}
\orcid{0009-0005-0585-1642}
\affiliation{
    \institution{Czech Technical University in Prague, Faculty of Electrical Engineering}
    \city{Prague}
    \country{Czech Republic}}
\email{contact@nemecekjiri.cz}

\author{Mark Kozdoba}
\orcid{0000-0002-8451-023X}
\affiliation{
    \institution{Technion}
    \city{Haifa}
    \country{Israel}}
\email{markk@technion.ac.il}
    
\author{Illia Kryvoviaz}
\orcid{0009-0002-9834-6722}
\affiliation{
    \institution{Czech Technical University in Prague, Faculty of Electrical Engineering}
    \city{Prague}
    \country{Czech Republic}}
\email{illiakryvoviaz@gmail.com}
    
\author{Tomáš Pevný}
\orcid{0000-0002-5768-9713}
\affiliation{
    \institution{Czech Technical University in Prague, Faculty of Electrical Engineering}
    \city{Prague}
    \country{Czech Republic}}
\email{pevnak@protonmail.ch}
    
\author{Jakub Mareček}
\orcid{0000-0003-0839-0691}
\affiliation{
    \institution{Czech Technical University in Prague, Faculty of Electrical Engineering}
    \city{Prague}
    \country{Czech Republic}}
\email{jakub.marecek@fel.cvut.cz}

\copyrightyear{2025}
\acmYear{2025}
\acmConference[KDD '25] {Proceedings of the 31st ACM SIGKDD Conference on Knowledge Discovery and Data Mining V.2}{August 3--7, 2025}{Toronto, ON, Canada}
\acmBooktitle{Proceedings of the 31st ACM SIGKDD Conference on Knowledge Discovery and Data Mining V.2 (KDD '25), August 3--7, 2025, Toronto, ON, Canada}
\acmDOI{10.1145/3711896.3736857}
\setcopyright{cc}
\setcctype{by}
\acmISBN{979-8-4007-1454-2/2025/08}

\renewcommand{\shortauthors}{Jiří Němeček, Mark Kozdoba, Illia Kryvoviaz, Tomáš Pevný, \& Jakub Mareček}

\title{Bias Detection via Maximum Subgroup Discrepancy}
\date{\today}

\begin{document}


\begin{abstract}
Bias evaluation is fundamental to trustworthy AI, both in terms of checking data quality and in terms of checking the outputs of AI systems. In testing data quality, for example, one may study the distance of a given dataset, viewed as a distribution, to a given ground-truth reference dataset.
However, classical metrics, such as the Total Variation and the Wasserstein distances, are known to have high sample complexities and, therefore, may fail to provide a meaningful distinction in many practical scenarios. 

In this paper, we propose a new notion of distance, the Maximum Subgroup Discrepancy (MSD). In this metric, two distributions are close if, roughly, discrepancies are low for all feature subgroups. While the number of subgroups may be exponential, we show that the sample complexity is linear in the number of features, thus making it feasible for practical applications. 
Moreover, we provide a practical algorithm for evaluating the distance based on Mixed-integer optimization (MIO). 
We also note that the proposed distance is easily interpretable, thus providing clearer paths to fixing the biases once they have been identified. 
Finally, we describe a natural general bias detection framework, termed MSDD distances, and show that MSD aligns well with this framework. 
We empirically evaluate MSD by comparing it with other metrics and by demonstrating the above properties of MSD on real-world datasets. 
\end{abstract}

\maketitle

\section{Introduction}
\label{sec:intro}

Regulatory frameworks, such as the AI Act \cite{AIAct} in Europe, suggest that one needs to measure data quality, including bias detection in training data,
as well as to detect bias in the output of the AI system, but provide no suggestions as to what bias measures to use.
This is the case of the very recent IEEE Standard for Algorithmic Bias Considerations \cite{10851955} and earlier NIST white papers \cite{schwartz2022towards}  too, where the latter stops at the ``majority of fairness metrics are observational as they can be expressed using probability statements involving the available random variables''.

At the most basic level, one could imagine bias detection as a two-sample problem in statistics, where, given two sets of samples, one asks whether they come from the same distribution.
In practice, the two sets of samples often do not come from the same distribution, but one would like to have an estimate of the distance between the two distributions.
The distance estimate, as any other statistical estimate \cite{tsybakov2009nonparametric}, comes with an error.
One would like the error in the estimate to be much smaller than the estimated value
for the bias detection to be credible, stand up in any court proceedings, etc.


\begin{figure}
    \centering
    \includegraphics[width=0.9\linewidth]{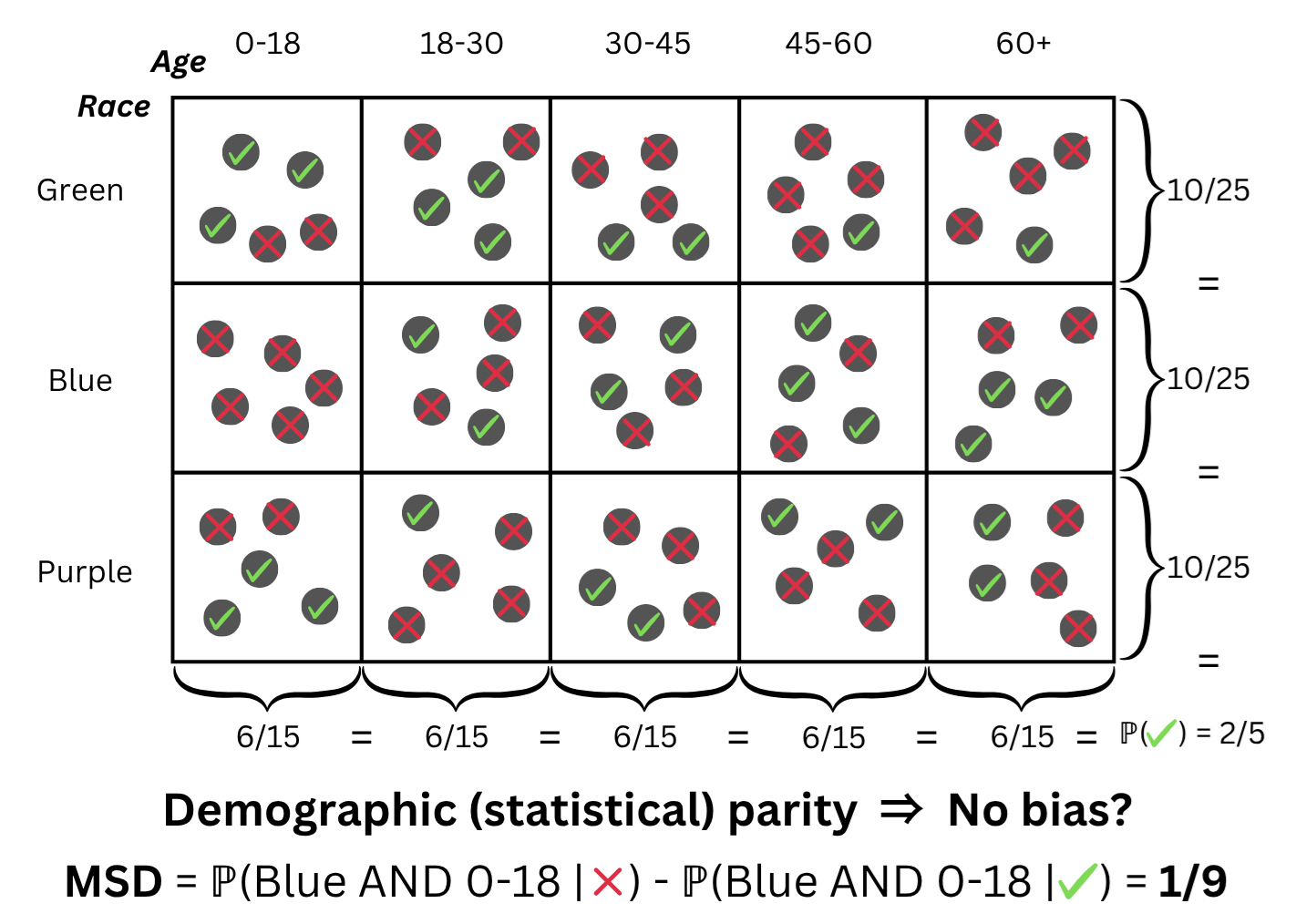}
    \caption{An illustration of intersectional bias. A classifier accepts an equal proportion of persons of each race and age group. Thus, demographic parity is achieved, and no such bias is detected. However, when considering subgroups, i.e., intersections of protected attributes, we see that young blue people are severely disadvantaged. This disparity is found by our proposed measure, Maximum Subgroup Discrepancy (MSD), which also recovers the description of the most disadvantaged subgroup (\texttt{race = "Blue"} AND \texttt{age $\in$ [0, 18]}).}
    \Description{A figure illustrating the importance of intersectional fairness. In two dimensions, age and race are binned into 3 by 5 bins, each having the same number of samples. Each sample is either classified as positive or negative. In each row and column is the same proportion of positives and negatives, which means that demographic parity is achieved. However, looking at the bin of young blue people, one notices that all of them are classified as negative. The proposed measure, MSD, is able to capture that and recover the most disadvantaged subgroup.}
    \label{fig:intersectional_fairness}
\end{figure}

In intersectional fairness, one would like to estimate bias for all protected subgroups. See Figure \ref{fig:intersectional_fairness} for a motivating toy example.
Estimation of intersectional bias faces three challenges:
(1) the number of subgroups is exponential in the number of protected attributes, 
(2) estimating many distances on measure spaces has sample complexity exponential in the ambient dimension for each subgroup, and
(3) some subgroups may have too few samples to estimate the bias well in terms of estimation error and significance. 
Let us now consider these challenges in more detail. 

Sample complexity is the number of samples that makes it possible to estimate a quantity to a given error.
A lower bound on sample complexity then suggests the largest known number of samples universally required to reach a given error.
The sample complexity of bias estimation depends on the distance between distributions (measures) used.
The accuracy improves with the number of samples taken,
but the rate of improvement depends on the dimension.
See Table \ref{tab:samplecomplexity} for a brief summary of the commonly considered distances and their sample complexity, and Section 
\ref{relateddistances} for references and discussion. 
As is often the case in high-dimensional probability, the ``curse of dimensionality'' suggests that the number of samples to a given error grows exponentially with the dimension.

Not only is sample complexity a lower bound on the runtime \cite{lee2023computability}, but one must often test for bias for all subgroups, of which there are exponentially many in the number of protected attributes.
Together, this could lead to doubly exponential time complexity, challenging even for low dimensions. Thus, one would like to detect bias for a single subgroup with polynomial complexity or reason about the joint problem.
Additionally, the number of samples representing a certain subgroup decreases with the number of protected attributes defining it. Smaller sample complexity enables us to reliably detect bias for smaller subgroups.

In this paper, we \textbf{(i)} introduce the Maximum Subgroup Discrepancy (MSD) metric. Compared to the standard distances discussed above,  we \textbf{(ii)} show that it has a manageable sample complexity (linear in the number of protected attributes). We also note that in contrast to other distances, the MSD is \emph{interpretable} and thus provides actionable information on how the bias in the data should be mitigated once found. It also provides a guarantee in the form of an upper bound on bias over all subgroups. Next, we \textbf{(iii)} develop a new mixed-integer optimization (MIO) formulation for the evaluation of $\MSDc$. Finally, we \textbf{(iv)} numerically validate the estimation stability and the dependence on the sample size for $\MSDc$ and demonstrate its advantage over common distribution distance measures on 10 real-world datasets, which were developed with the view of studying fairness.
We show that $\MSDc$ requires exponentially fewer samples to obtain a good estimate of the true value of the distance.
Taken together, we believe that these contributions 
suggest MSD as a theoretically grounded and practically useful bias detection method.

\subsection{Problem description}
\label{sec:problem_statement}
To arrive at the definition of MSD, we first discuss the most general intersectional bias problem.  
Suppose we are given two probability distributions $\posdistr$ and $\negdistr$ over an input space $\mcX \subseteq \RR^\ndims$. 
A subset of the features, denoted $\prot \subseteq \dims$,  will be considered to be the \emph{protected attributes}, 
such as for instance age, gender, ethnicity, and education. 
A \emph{subgroup} is a subset of a population with a given fixed value of one or more protected attributes (e.g., black women). 
Formally, a subgroup $\subg$, defined by attributes $\prot$ and their values $V=\Set{v_p}_{p \in \prot} \subseteq \RR^{\Abs{\prot}}$, is the set 
\begin{equation}
    S = S_{\prot,V} = \Set{x = (x_1, \ldots, x_d) \in \mcX \setsep x_p = v_p \text{ for $p \in \prot$}}.
\end{equation}
Denote by $\subgs$ the set of all possible subgroups $S$.
To study intersectional fairness, one would like to estimate the distance between the two distributions 
restricted to 
a given subgroup and find the subgroup for which these distributions differ most. Specifically, we call the Maximum Subgroup Distribution Distance
\begin{equation}
\mathrm{MSDD}_\distance(\posdistr, \negdistr; \prot) = \sup_{{\subg \in \subgs}} \distance({\posdistr_{|\subg},  \negdistr_{|\subg}}),
\end{equation}
where 
$\posdistr_{|\subg}$ defines the \emph{restriction} measure to the subset $\subg$: $\posdistr_{|\subg}(A) := \posdistr(A \cap \subg) / \posdistr(\subg)$,
and $\distance$ a (possibly semi-) metric on measure spaces over the input space, $\distance: \mathcal{M}(\mcX) \times \mathcal{M}(\mcX) \to \RR$. 
For instance, $\Delta$ could be the Total Variation or the Wasserstein distances (Section \ref{relateddistances}).

To give an example for the use of MSDD in evaluating data quality, $\posdistr$ could be training data, seen as an empirical distribution, and $\negdistr$ could be a distribution based on census data. Alternatively, we can consider distributions of a target variable of an ML model in auditing its fairness: $\posdistr$ for the positive class and $\negdistr$ for the negative. We can then find a subgroup that is most disadvantaged (in terms of $\mathrm{MSDD}_\distance(\posdistr, \negdistr; \prot)$) by the model.

The MSDD is a natural general metric that may be used to quantify intersectional bias in a variety of scenarios, including both data quality and system output settings. 
However, it also has a number of fundamental practical issues:
(i) The enumeration of all subgroups would have exponential time complexity in the number of protected attributes, and (ii) common distances on measure spaces ($\Delta$) have high sample complexity (Section \ref{relateddistances}), which makes them impossible to estimate for small subgroups. Note, however, that since the number of subgroups is large, some of the subgroups will almost \emph{inevitably} be small. 
(iii) The run time complexity of many standard distribution metrics is superlinear \cite{NEURIPS2022_f02f1185,gretton2012kernel}
which is then compounded by the need to compute the distances for many subgroups. 

To alleviate these issues, we introduce the MSD distance. The MSD can be used as the $\Delta$ for the $\mathrm{MSDD}$ distance and is itself based on the notion of subgroups (the attributes $\prot$ along which $\mathrm{MSDD}$ is computed would normally be disjoint from the ones used to define 
MSD). As discussed earlier, MSD has favorable sample complexity properties and an efficient computation algorithm, thus providing considerable stability and speed gains for issues (ii) and (iii). 

The rest of this paper is organized as follows:
The background and prior work are discussed in Section 
 \ref{sec:background}. In Section \ref{sec:max_norm} we introduce the MSD, discuss a number of its properties, and prove the sample complexity bounds.  In Section \ref{sec:estimation} we introduce the MSD estimation algorithm, and Section \ref{sec:experiments} is dedicated to the empirical evaluation. Finally, we conclude the paper in Section \ref{sec:conclusion}.

\section{Background and Related work}
\label{sec:background}
\begin{table}
    \centering
    \caption{ 
    Sample complexity of estimating popular distances on measure spaces in terms of the ambient dimension $\ndims$.
    }
    \resizebox{\linewidth}{!}{
    \begin{tabular}{lcc}
        \toprule
        Distance & Samples & Reference \\
        \midrule
        Wasserstein-1 & $\infty$ & Thm 5 \cite{lee2023computability} \\ 
           Total Variation & $\bigomega{\exp(d)}$ & \cite{devroye2018total,wolfer2021statistical} \\ 
        Hellinger (Jeffreys)  & $\bigomega{\exp(d)}$ & \cite{devroye2018total,wolfer2021statistical} \\     
        Wasserstein-2 & $\bigomega{\exp(d)}$ & \cite{dudley1969speed,fournier2015rate,weed2019sharp} \\ 
        Wasserstein-$\infty$ & $\bigomega{\exp(d)}$ & \cite{fournier2015rate,liu2018rate,weed2019sharp} \\ 
        MMD & $\bigomega{\exp(d)}$ & \cite{NIPS2016_5055cbf4} \\
        \midrule
        $\MSDc$ & $\bigoh{\ndims}$ w.h.p. & This work (Eq. \ref{eq:smpl_thm_pr2} in Sec. \ref{sec:complexity}) \\
        \bottomrule
    \end{tabular}}
    \label{tab:samplecomplexity}
\end{table}

\subsection{Distances on Measure Spaces}
\label{relateddistances}

There are numerous distances on measure spaces used in applied probability, including (in the approximately chronological order)
Total Variation (TV, \cite{zbMATH02706589}), 
Hellinger distance \cite{hellinger1909neue},
Kullback–Leibler (KL) divergence \cite{KL} and its variants, 
Wasserstein-2 \cite{vaserstein1969markov,dudley1969speed} and its variants such as Wasserstein-1 \cite{vaserstein1969markov},  and 
Maximum Mean Discrepancy (MMD, \cite{gretton2012kernel}).
As we outline below,   most of these methods have exponential sample complexity in terms of the number of dimensions.
As can be seen in Table~\ref{tab:samplecomplexity}, most of these distances come with either undecidability results \cite{lee2023computability} or 
exponential lower bounds on their sample complexity in the worst case. 
For MMD, while \cite{gretton2012kernel} claimed polynomial sample complexity,  \cite{NIPS2016_5055cbf4} explained the lower bounds on sample complexity under strong assumptions.
Yet for other methods, such as Hellinger and Jeffreys distances, 
high sample complexity follows from their relationship to TV distance.
We refer to \cite{panaretos2019statistical} for a thorough survey.

Obviously, one can consider additional assumptions, such as having cardinality of the support (which scales with $\exp(d)$ in general), bounded by a constant. Testing TV closeness in time sublinear in the support was derived in
\cite{chan2014optimal}. 
Entropy estimation bounds were obtained in \cite{valiant2011estimating}.
\cite{feng2023simple,bhattacharyya2022approximating} give an algorithm for estimating TV between just \emph{product} measures, polynomial in product dimension.
Likewise, one can consider smoothness of the measures and certain invariance properties \cite{chen2023sample,tahmasebisample}, or focus on Gaussian distributions \cite{hsu2024polynomial} only.
Ising-type models testing was explored in \cite{kandiros2023learning}. While these assumptions are of considerable interest, it is not easy to test whether those assumptions hold in real-world data sets.


\subsection{Subgroup and Intersectional Fairness}

The notion of subgroups gave rise to the 
 work on subgroup fairness \cite{kearns2018preventing}, and underlies the work on intersectional fairness \cite{foulds2020intersectional,gohar2023survey}. 
In particular, in the legal scholarship, Intersectional Fairness ideas go back to the work of 
Crenshaw \cite{crenshaw2013demarginalizing}, but remain a subject of lively debate \cite{collins2020intersectionality} to the present day. 
In the algorithmic fairness literature, the sample complexity of certain fairness estimates (statistical parity, false positive fairness) was considered in \cite{kearns2018preventing}. The notion of distance, which we consider here, is conceptually different, as it concerns data quality (see Section \ref{sec:intro}) rather than a fairness test of a particular given classifier. Moreover, 
the algorithms of \cite{kearns2018preventing} were developed with \emph{linear} subgroups in mind, and consequently evaluated on linear subgroups \cite{kearns2019empirical}. 
In particular, these algorithms require certain specific heuristics (oracles) that are mainly developed for the linear case. 
We note that such linear subgroups are considerably less interpretable and less suitable for real-world applications compared to the intersectional subgroups that we consider here.  
Note that naively extending the definitions used in marginal fairness (considering only groups defined by a single protected attribute at a time \citep{10420507}) to intersectional fairness will suffer from low sample sizes of small subgroups. Estimates using such extensions \citep{10.1145/3597503.3639083} will have dubious significance and will be less robust to outliers. 

In a related direction, multidimensional subset scanning methods systematically sift through potentially large collections of subgroups to find anomalous or biased regions. Although these approaches vary in the exact objective -- ranging from pinpointing classifier miscalibration \cite{zhang2016identifying} to detecting compact anomalies via penalties \cite{speakman2015penalized} or scanning multiple data streams \cite{neil2013fast} -- they share a focus on likelihood-based scoring and efficient ``fast subset'' searches. While these techniques are adept at finding one or more ``highest-scoring'' subgroups, their goals and scoring mechanisms are distinct from distance-based comparisons.

Finally, there is vast literature on the topic of debiased learning \citep[e.g.,][]{10.5555/3524938.3524988,NEURIPS2020_eddc3427,ahn2023mitigating}, which is concerned with learning unbiased models on biased data. Said bias is, however, in the form of spurious correlations, making the task distinct from the one considered here. 


\subsection{Learning DNF}
A (protected) subgroup is naturally defined as a conjunction of a few feature-value pairs, e.g., \texttt{sex = "F" AND race = "black"}. A disjunction of multiple conjunctions (i.e., a union of subgroups) is a logical formula in Disjunctive Normal Form (DNF).

The study of learning DNF formulas is a fundamental part of theoretical computer science. Despite its polynomial sample complexity, finding a general algorithm with polynomial-time complexity has proven elusive in many settings.
Since the breakthrough reduction to polynomial threshold functions of \cite{klivans2001learning}, which matched the 1968 lower bound of \cite{minsky1988perceptrons},  
 \cite{daniely2016complexity} helped understand the complexity-theoretic barriers.
See also \cite{shalev2014understanding} for an overview.

\subsection{Mixed-integer optimization}
Mixed-integer optimization (MIO, \cite{wolseyIntegerProgramming2021}) is a powerful framework for modeling and solving mathematical optimization problems, where some decision variables take values from a discrete set while others are continuously valued. 
Despite MIO being a general framework for solving NP-hard problems, the MIO solvers are speeding up by approximately 22\% every year, \emph{excluding} hardware speedups \cite{kochProgressMathematicalProgramming2022}.  
We use the abbreviation MIO, though we consider only mixed-integer \emph{linear} formulations.

MIO is used in machine learning, especially when one optimizes over discrete measures or decisions. This includes learning logical rules, including DNFs. \citet{malioutovExactRuleLearning2013} learn DNFs through a sequential generation of terms using an LP relaxation of an MIO formulation. Later, \citet{wangLearningOptimizedOrs2015} used MIO to optimize full DNFs, and \citet{suLearningSparseTwolevel2016} introduced a formulation with Hamming distance as an alternative objective function. A crucial improvement to the scalability of exact learning of DNFs was the BRCG \cite{dash2018boolean}, which utilizes column generation to generate candidate terms. Recently, MIO was utilized to learn a DNF classifier with fairness constraints \cite{lawlessInterpretableFairBoolean2023}. Importantly, we optimize only a single term (conjunction), representing the maximally discrepant subgroup, to evaluate bias. We do not perform predictions.







As an aside, note that \cite{nair2021changed} (see also \cite{haldar2023interpretable}) uses DNFs to compare two \emph{models}.
This is done by building two DNFs as proxy models and comparing them. 
However, we could also directly compare the distributions of the two models, as proposed here. 

\section{Maximum Subgroup Discrepancy}
\label{sec:max_norm}

Let $\mcX \subset \RR^d$ be an input space with $\ndims$ features, and let $\prot \subseteq \dims$ be a subset of features that are protected attributes. Throughout the rest of the paper, we assume that the protected attributes are binary, which can always be achieved by quantization and one-hot encodings. 
For a vector $x\in \RR^d$ and 
$p \in \prot$ let 
$x_p \in \Set{0,1}$ denote its $p$-th coordinate, 
and let $\bar{x}_p := 1-x_p$ denote its logical negation. 
The set of functions $\mcL = \Set{x_p,\bar{x}_p}_{p \in \prot}$ is called \emph{literals}. 
A \emph{term} is a conjunction of literals. That is, for a subset $S \subset \mcL$, the term $\chi_S$ is defined as a function $\RR^d \rightarrow \Set{0,1}$ given by 
\begin{equation}
    \chi_S(x) = \prod_{s \in S} s(x). 
\end{equation}
The subsets $S \subset \mcL$ are called \emph{subgroups}, and with some abuse of notations, we also refer to the corresponding parts of the data defined by $S$, 
\begin{equation}
    \Xsubg = \Set{x \in \RR^d \setsep \chi_S(x) = 1},
\end{equation}
as subgroups. 
The set of all possible subgroups $S \subset \mcL$ is denoted by $\subgs$.

For a distribution $\mu$ on $\RR^d$ and a function $f:\RR^d \rightarrow \RR$ denote by $\mu(f)$ the integral $\mu(f) = \int f(x) d\mu(x)$. 

With this notation, the $\MSDc$ distance on feature set $\prot$ between two distributions, $\mu,\nu$, is defined as 
\begin{equation}
\label{eq:main_msd_def}
    \MSDc(\mu,\nu;\prot) = \sup_{S \in \subgs} \Abs{\mu(\chi_S) - \nu(\chi_S)}. 
\end{equation}
In words, as discussed earlier, two distributions are similar w.r.t MSD if all subgroups induced by $\prot$ have similar weights in both measures.  

Let us now discuss the relation between MSD and the two standard distances - the $\ell_{\infty}$ and $\ell_1$, or equivalently, the TV distance. 
Define the base terms as 
\begin{equation}
    \mcB_{\prot} = \Set{ \prod_{p\in \prot} l_p \setsep \text{ where $l_p = x_p$ or $l_p = \bar{x}_p$} } \subset \subgs.
\end{equation}
That is, the terms in $\mcB_{\prot}$ correspond to all possible different values the projection of $x$ onto features $\prot$ might have. Equivalently, base terms correspond
to subgroups where \emph{all} protected attributes have a specified value (rather than the more general specification of only part of the values).  Clearly, there are precisely 
$2^{\Abs{\prot}}$ base terms. 

The $\ell_{\infty}$ norm may then be defined as 
\begin{equation}
\label{eq:sup_def}
    \norm{\mu - \nu}_{\infty} = \sup_{S \in \mcB} \Abs{\mu(\chi_S) - \nu(\chi_S)}. 
\end{equation}
That is, we consider the maximal weight difference over the atoms in $\prot$. 

Let $\mcF_{\prot,\infty}$ be the set of functions $f: \RR^d \rightarrow \RR$ that depend only on the coordinates in $\prot$, and are bounded by $1$. Then, the \emph{total variation}\footnote{marginalised to $\prot$} (TV),  may be written as 
\begin{flalign}
    \label{eq:tv_line1}
    \norm{\mu - \nu}_{TV} &= \half \sup_{f \in \mcF_{\prot,\infty}}
    \Abs{\mu(f) - \nu(f)}
    \\    
    \label{eq:tv_line2}
    &= \half \sum_{S \in \mcB} \Abs{\mu(\chi_S) - \nu(\chi_S)},
\end{flalign}
where \eqref{eq:tv_line1} and \eqref{eq:tv_line2} are the dual and the standard definitions of the $\ell_1$ (and hence the TV) norms. 

By comparing the definition \eqref{eq:main_msd_def} with \eqref{eq:sup_def} and \eqref{eq:tv_line1}, we see that 
\begin{equation}
    \norm{\mu-\nu}_{\infty} \leq \MSDc(\mu,\nu;\prot) \leq 
    \norm{\mu-\nu}_{TV},
\end{equation}
where the second inequality follows since all terms in $\subgs$ are clearly functions bounded by 1. 

We thus observe that $\MSDc(\mu,\nu;\prot)$ is a stronger distance than $\ell_{\infty}$, which allows us to consider subgroups with partially specified values. At the same time, it is not as strong as the total variation, which requires the \emph{sum} over all base term differences to be small (rather than each of the differences being small individually). On the other hand, as discussed in Section \ref{sec:intro}, TV is, in fact, too strong to be practically computable due to its high sample complexity, while $\MSDc$ has a sample complexity linear in the number of attributes.

\subsection{Sample Complexity}
\label{sec:complexity}

In this section, we prove Theorem \ref{thm:main_sample_thm}, 
which allows us to quantify the error of estimating 
$\MSDc(\mu,\nu;\prot)$ from finite samples of these distributions. 

Let $\Set{x_i}_{i\leq N_1}$ be $N_1$ independent samples from $\mu$, and $\Set{x'_i}_{i\leq N_2}$ be $N_2$ independent samples from $\nu$.  Define the corresponding \emph{empirical distributions}, $\hat{\mu},\hat{\nu}$ by 
\begin{equation}
    \hat{\mu} = \frac{1}{N_1} \sum_{i} \delta_{x_{i}} 
    \text{ and } \hat{\nu} = \frac{1}{N_2} \sum_{i} \delta_{x'_{i}},
\end{equation}
where $\delta_{x}$ is an atomic distribution at point $x$, with weight $1$.

\begin{theorem} 
\label{thm:main_sample_thm}
Fix $\delta>0$ and set $N = \min(N_1,N_2)$. Then with probability at least $1-2\delta$ over the samples,  
\begin{equation}
    \MSDc(\mu,\nu;\prot) \leq \MSDc(\hat{\mu},\hat{\nu};\prot) + 
    4\sqrt{\frac{2\Abs{\prot} + \log \frac{2}{\delta}}{2N}}.  
\end{equation}
\end{theorem}
In words, if the number of samples $N$ is of the order of the number of protected attributes, $\Abs{\prot}$, or larger, we can estimate $\MSDc(\mu,\nu;\prot)$ by computing the $\MSDc$ on the empirical distributions, $\MSDc(\hat{\mu},\hat{\nu};\prot)$.
\begin{proof}
For any term $S \in \subgs$, by the triangle inequality we have 
\begin{flalign*}
\Abs{\mu(S) - \nu(S)} \leq 
&\Abs{\hat{\mu}(S) - \hat{\nu}(S) } \\ 
&+ 
\Abs{\mu(S) - \hat{\mu}(S) } + 
\Abs{\nu(S) - \hat{\nu}(S) }.
\end{flalign*}
Therefore, taking suprema over $S$, we have 
\begin{flalign*}
\label{eq:smpl_thm_pr1}
\MSDc(\mu,\nu;\prot) \leq \; &  \MSDc(\hat{\mu},\hat{\nu};\prot) \\
&+ 
\sup_{S\in \subgs} \Abs{\hat{\mu}(S) - \mu(S)} + 
\sup_{S\in \subgs} \Abs{\hat{\nu}(S) - \nu(S)}.    
\end{flalign*}
The quantity $\sup_{S\in \subgs} \Abs{\hat{\mu}(S) - \mu(S)}$
describes the deviation of the empirical mean from the true mean over the $\subgs$  family of functions. Therefore, 
by the uniform concentration results for bounded finite families (see, for instance, Theorem 2.13 in \cite{mohri2018foundations}), with probability at least $1-\delta$ over $\Set{x_i}$, we have 
\begin{equation}
\label{eq:smpl_thm_pr2}
\sup_{S\in \subgs} \Abs{\hat{\mu}(S) - \mu(S)} \leq 
2\sqrt{\frac{\log \Abs{\subgs} + \log \frac{2}{\delta}}{2N_1}}.
\end{equation}

Next, to estimate $\Abs{\subgs}$, observe that for every $p\in \prot$, a term either contains $x_p$, or $\bar{x}_p$, or neither. Thus there are at most $3^{\Abs{\prot}}$ terms. In particular, we have 
\begin{equation}
\label{eq:smpl_thm_pr3}
    \log \Abs{\subgs} \leq 2 \Abs{\prot}.
\end{equation}

Finally, the proof is completed by repeating the argument for $\nu$, combining the results above, and using the union bound. 
\end{proof}

\subsection{MSD Estimation As Classification}
\label{sec:MSDclassification}
To estimate the $\MSDc$ for empirical distributions $\hat{\mu},\hat{\nu}$, it is more convenient to rephrase the maximization problem \eqref{eq:main_msd_def} as minimization of a classification loss, where the classifiers are taken from the family of terms, $\subgs$. As we detail in Section \ref{sec:estimation}, this point of view allows us to incorporate various useful ideas from the field of DNF classification into our MIO-based minimization algorithm.

To recast the problem as classification, denote by $\Set{x_i}_{i\leq N_1}$ and the $\Set{x'_j}_{j\leq N_2}$ datasets sampled from $\mu$ and $\nu$ respectively. Assign a label $y=1$ to all samples from $\mu$ and $y=0$ to all samples from $\nu$. 
Let $c(a,b)$ be the binary classification loss (0-1 loss), 
$c(a,b) = 0 \text{ if } a=b \text{ and } 1 \text{ otherwise}$. 

Then, for a binary classifier $f$ of a label $y$ as above, the binary classification loss would be 
\begin{flalign*}
    &\Expsubidx{x\sim \mu} c(f(x),1) + \Expsubidx{x\sim \nu} c(f(x),0) \\
    &=  \Expsubidx{x\sim \mu} \Brack{1-f(x)} + 
    \Expsubidx{x\sim \nu} f(x)  \\ 
    &= 1 -\Expsubidx{x\sim \mu} f(x)
    +\Expsubidx{x\sim \nu} f(x) \\ 
    &= 1 - \Brack{ \mu(f) - \nu(f)}
\end{flalign*}
Thus, minimizing the classification loss is equivalent to maximizing the integral difference $\mu(f) - \nu(f)$, on which MSD is based. 
Minimizing the classification loss with flipped labels (i.e., $y^\prime = 1 - y$) is equivalent to maximizing the opposite difference ($\nu(f) - \mu(f)$), enabling us to find the maximal absolute difference.

Note that the role of the classifier here is different from that of typical classifiers in algorithmic fairness, as $f$ is not required to be fair in any way. Rather, its role is similar to an \emph{adversary} in adversarial machine learning and is required to distinguish the two datasets as well as possible.

\section{Practical Estimation}
\label{sec:estimation}

In Section \ref{sec:max_norm}, we have described a measure with practically feasible sample complexity requirements. However, there is still a question of how to estimate it in practice. The challenge lies in the cardinality of $\subgs$, which is exponential in the number of protected attributes. See Figure \ref{fig:enumerative} for an illustration.

Thus, there is a question of how to estimate the full $\MSDc$ distance effectively. To this end, recall that subgroups $\subg \in \subgs$ constitute \emph{terms}, i.e., conjunctions of literals formed on a subset of features $\prot$.
We also showed that one can formulate the $\MSDc$ distance maximization using two classification problems. 

By using the methods of \cite{wangLearningOptimizedOrs2015,arya2019one,wei2019generalized}, we could learn a DNF that distinguishes between $\posdistr$ and $\negdistr$. Note that by definition, every term in such a DNF would correspond to some subgroup $\subg$. 
Since we search for a single member $\subg \in \subgs$, we do not need to search for a full DNF, we rather need a single term -- single conjunction.  

To ensure that by minimizing the 0-1 loss, we indeed optimize the same notion as searching for the subgroup with the maximal discrepancy, we must have balanced classes, i.e., an equal number of samples for each label. Alternatively, we can weigh the 0-1 losses of samples from a given class by $1/\text{number of samples of a given class}$. 
Thus, we can run any conjunction-learning algorithm to obtain a subgroup and the loss $L_1$, flip the labels $y_i$, and optimize again to get the loss $L_2$. Then 
$\MSD[]{\hat{\posdistr}}{\hat{\negdistr}}\prot = 1 - \min(L_1, L_2)$, 
assuming the found solutions of the classification problems are globally optimal, i.e., the found subgroups (terms) have the minimal loss out of all in $\subgs$.

\subsection{MIO formulation}

To find the globally optimal subgroup (i.e., conjunction or 1-term DNF), one can utilize Mixed-Integer Optimization (MIO). Let $\datapos$ and $\dataneg$ be sets of indices of samples from $\posdistr$ and $\negdistr$, respectively, and let $\data = \datapos \cup \dataneg$ be the set of indices of all samples.
Our formulation is related to the 0-1 error DNF formulation of \cite{suLearningSparseTwolevel2016}, but we search for a single term only and use a different objective, which leads to us also having to use further constraints.

Since we maximize the non-linear absolute difference, we must reformulate it using an auxiliary variable. We define variable $o$ as the absolute value objective we maximize. We bound it from above by the two potential values of the absolute value, such that it takes the higher of them.  The entire formulation is 
\begin{subequations}
\label{eq:mio}
\begin{align}
    \max \; & o \\
    \mathrm{s.t.} \; & o \le \frac{1}{\Abs{\datapos}} \sum_{i \in \datapos}{\yhat} - \frac{1}{\Abs{\dataneg}} \sum_{i \in \dataneg}{\yhat} + 2b \label{eq:mio_abs1} \\
    & o \le \frac{1}{\Abs{\dataneg}} \sum_{i \in \dataneg}{\yhat} - \frac{1}{\Abs{\datapos}} \sum_{i \in \datapos}{\yhat} + 2(1-b) \label{eq:mio_abs2}  \\
    & \yhat \le 1 - (\usevar - x_{i,j} \cdot \usevar) \hspace{1cm} i \in \data, \; j \in \prot \label{eq:mio_positive} \\
    & \yhat \ge 1 - \sum_{j \in \prot}(\usevar - x_{i,j} \usevar) \hspace{1.8cm} i \in \data \label{eq:mio_negative} \\
    & \sum_{i \in \data} \yhat \ge \minSize &  
    \label{eq:mio_lowerbound} \\
    & 0 \le \yhat \le 1 \hspace{4.1cm}  i \in \data \label{eq:mio_yhatbounds}\\
    & \usevar, b \in \{0, 1\} \hspace{3.7cm} j \in \prot,
\end{align}
\end{subequations}
where $\usevar$ is equal to 1 if and only if feature $j$ is present in the conjunction, and $\yhat \in \{0,1\}$ are variables representing whether each sample belongs to the subgroup (i.e., is classified as 1). Indeed, $\yhat$ takes only binary values, due to the constraints on its value, together with the fact that $x_{i,j}$ and $\usevar$ are binary.

The first two constraints---\eqref{eq:mio_abs1} and \eqref{eq:mio_abs2}---formulate the absolute value objective. The binary variable $b$ serves to relax one of the upper bounds to allow the objective variable $o$ to be constrained by the higher bound out of the two. The value 2 is a tight ``big-M'' value in this context since the difference of the two means takes values from $[-1, 1]$.  

Third constraint \eqref{eq:mio_positive} forces $\yhat = 0$ for samples that have $x_{i,j} = 0$ but $\usevar = 1$ for some feature $j \in \prot$, i.e., when the literal for sample $x_i$ on feature $j$ is not satisfied. Since this is a conjunction, a single such event means that the sample will not belong to the subgroup (i.e., will be classified as 0).
The fourth constraint \eqref{eq:mio_negative} ensures $\yhat = 1$ when all literals are satisfied by the sample $x_i$. 

Finally, the last constraint \eqref{eq:mio_lowerbound} introduces a lower bound on the number of samples that form a feasible subgroup. This can be used to restrict the search to subgroups that are reliably represented in the sampled data.  

If all $\usevar = 0$, the formulation represents an empty conjunction that always returns true, i.e., $\yhat = 1$ for all $i \in \data$. This represents the trivial subgroup containing all samples. 

By solving this MIO problem, we find the subgroup with the highest absolute difference in probability of belonging to $\posdistr$ or $\negdistr$, i.e., the empirical estimate of $\MSDdiff$.

\section{Experiments}
\label{sec:experiments}
Experimentally, we aim to answer the following questions:
\begin{itemize}
\item[Q1] Can the use of mixed-integer programming in \eqref{eq:mio} 
 be efficient? What is the run time in practice? 
\item[Q2] Does $\MSDc$ have a low sample complexity, as predicted by Theorem \ref{thm:main_sample_thm}?  
\item[Q3] Could one use alternative algorithms for learning DNF in estimating $\MSDc$?
\end{itemize}

Notice that the answers to all three questions are non-obvious. In Q1, we wish to compare different algorithms for evaluating MSD, whose run-time is exponential in the number of protected attributes.

In Q2, we want to corroborate our sample complexity results of Section \ref{sec:complexity}. We thus wish to compare $\MSDc$ to other measures of distances on distributions, as per Table \ref{tab:samplecomplexity}.

In Q3, we would like to compare the evaluation of $\MSDc$ utilizing the proposed MIO formulation to other DNF learning algorithms, even if not utilized in bias detection previously. We choose the classical algorithm Ripper \cite{ripper} and the more modern BRCG \cite{dash2018boolean}.

\subsection{Datasests}
To showcase the intended use of our method, we use the US Census data via the folktables library \cite{ding2021retiring}. Specifically, we use two families of five datasets. 


The first family consists of the five pre-defined prediction tasks provided by the folktables library \cite{ding2021retiring}. We compare the distributions of samples with $y_i = 1$ and $y_i = 0$ on the data of Californians in 2018. For example, in ACSIncome, this means comparing the distributions between people earning more than \$50,000 a year and people earning less, similar to the well-known Adult dataset. For more details, see the original paper \cite{ding2021retiring}. To make the distance evaluation comparable between methods, we only consider protected attributes. Those are 
listed in Appendix \ref{app:protected}. 

The second family of datasets aims to compare population distributions between different US states. We take 14 attributes that could be considered protected (e.g., sex, race, disability, age) and take data for five distinct pairs of states.
The state selection aims to compare demographically, socioeconomically, and culturally different states, allowing for a comparison of diverse populations across the United States. 

The sample sizes span from around 14,000 to 380,000, and the number of protected features spans from 4 (with around 1,200 possible subgroups) to 14 (with more than $5 \cdot 10^9$ subgroups). 
See Table \ref{tab:datasets} in Appendix \ref{app:data} for more details. 

\subsection{Setup}
To solve the proposed MIO formulation \eqref{eq:mio}, we model it using the Pyomo library \cite{bynumPyomoOptimizationModeling2021}, enabling easy use of a variety of solvers. We use the Gurobi solver \cite{gurobi}. We set the $\minSize$ parameter to 10, allowing only subgroups with at least ten samples.

We also implement TV and MMD. 
In the MMD evaluation, we utilize the overlap kernel function $k(x_{i_1},x_{i_2}) = \sum_{j = 1}^{\ndims} \istrue{x_{i_1,j} = x_{i_2,j}} / \ndims$ because the considered data is binary or categorical. For W$_1$ and W$_2$, we utilize POT \cite{flamary2021pot}, an open-source Python library for the computation of optimal transport.

Since there is no public implementation of the original BRCG, we take the non-MIO re-implementation of BRCG, called BRCG-light \cite{arya2019one}. We use the Ripper algorithm available in the AIX360 library \cite{arya2019one}. We modify both implementations to return a single-term DNF (i.e., a conjunction) to ensure comparability to our method. 
As addressed earlier, we run Ripper and BRCG two times, once to minimize error and the second time to maximize it (minimize the error with flipped labels). If one of the distributions has more samples, we subsample it to have an equal number of samples for each distribution. This makes minimizing the 0-1 loss equivalent to finding the $\MSDc$.

Each dataset is subsampled to form 5 (smaller) datasets, with sizes forming a geometric sequence from 1000 samples to the maximum number of samples for the given dataset. 
Each method is evaluated on each subsampled dataset. 
We ran each experiment configuration five times with different random seeds.
Each distance computation had a time limit of 10 minutes. 
All code for replicating the results is openly available on GitHub\footnote{\url{https://github.com/Epanemu/MSD}}. MSD is also implemented as a part of \texttt{humancompatible.detect} package\footnote{\url{https://github.com/humancompatible/detect}}.

\subsection{Results}
We evaluate three criteria. Firstly, we validate that our approach using MIO improves on naive enumeration of all subgroups (Q1). Secondly, we show that our approach indeed has advantageous sample complexity compared to other distribution distance measures (Q2). Finally, we validate that formulating the $\MSDc$ within MIO outperforms standard algorithms for finding DNFs when utilized to calculate $\MSDc$ (Q3) and showcase the interpretability of the proposed measure.

\subsubsection{Evaluating MS(D)D by enumeration (Q1)}
\begin{figure}[t]
    \centering
    \includegraphics[width=0.8\linewidth]{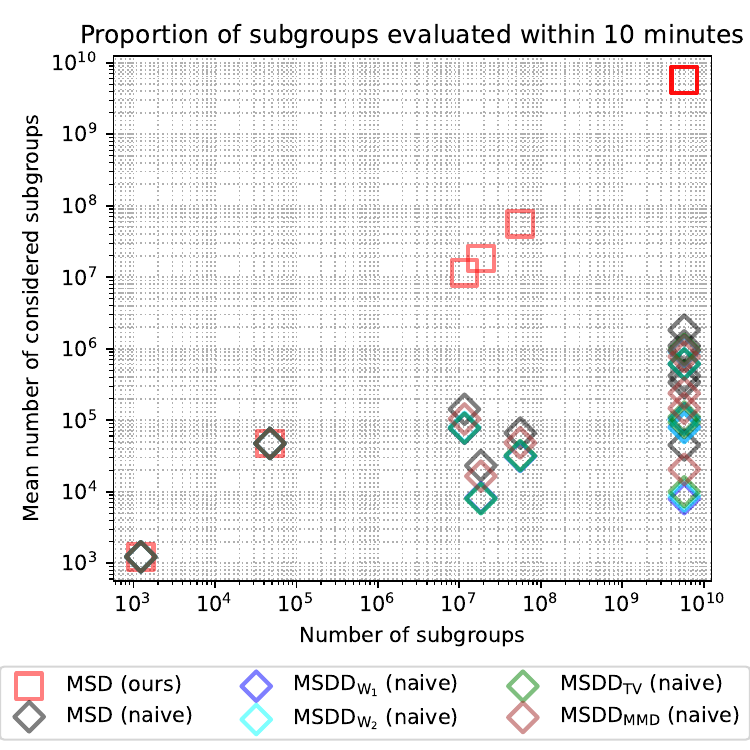}
    \caption{We show the number of subgroups in real datasets and the mean number of subgroups (subsets of the datasets) for which each distance was computed within 10 minutes. The diamonds represent measures computed naively by enumeration, and the squares represent the proposed method.
    }
    \Description{
    A log-log plot showing the number of subgroups on the x-axis and the number of considered subgroups within a 10-minute run time on the y-axis. A subgroup is considered if the distance on it was computed or if it has fewer than 10 samples. The proposed method is on the diagonal and dominates other methods that do not reach above 10 to the 7th power threshold.
    }
    \label{fig:enumerative}
\end{figure}

Naively, one could evaluate MSD (or MSDD, recall from Section \ref{sec:problem_statement}) by evaluating some distance for each of the exponential number of subgroups. We performed such an experiment to empirically compare this naive approach to the exponential runtime of solving the formulation \eqref{eq:mio}. 
We compare to a naive enumeration algorithm to compute $\MSDc$ and a few MSDD methods. They systematically enumerate subgroups and compute the given distance function $\Delta$ for each of them, seeking the maximum. We skip all encountered subgroups with fewer than 10 samples to ensure comparability. Subgroups with samples from only one distribution are also skipped when the distance cannot be computed. The enumeration was run on the same hardware, on a single thread in Python. The proposed method is capable of considering exponentially more subgroups in the evaluation of MSD, compared to the naive approach, when given the same time limit; see Figure \ref{fig:enumerative}. The $\MSDc$ points are on the diagonal since we know we have the maximum over \emph{all} subgroups when the formulation is solved to global optimality. We omit configurations that were not solved to optimality. 

For the datasets with the most subgroups, even the fastest method manages to consider around 3,000 times fewer subgroups. Extrapolating that, one would need more than 20 days to finish the computation. That is computed on the smallest dataset (Hawaii X Maine), where more than 89\% of the ``considered'' subgroups were skipped due to having too few samples.

\subsubsection{Evaluating sample complexity (Q2)}
Since most evaluated methods have different notions of distance, it is difficult to compare them directly. Thus, we present relative distances with respect to the best estimate of the distance. In other words, we divide each of the values by the mean distance computed on the entire dataset. Or, if that value cannot be computed, we take the mean value from the second-largest number of samples. For original distance measure comparison, see Figure \ref{fig:base} in Appendix \ref{app:base_distances}.

\begin{figure*}
    \centering
    \includegraphics[width=\textwidth]{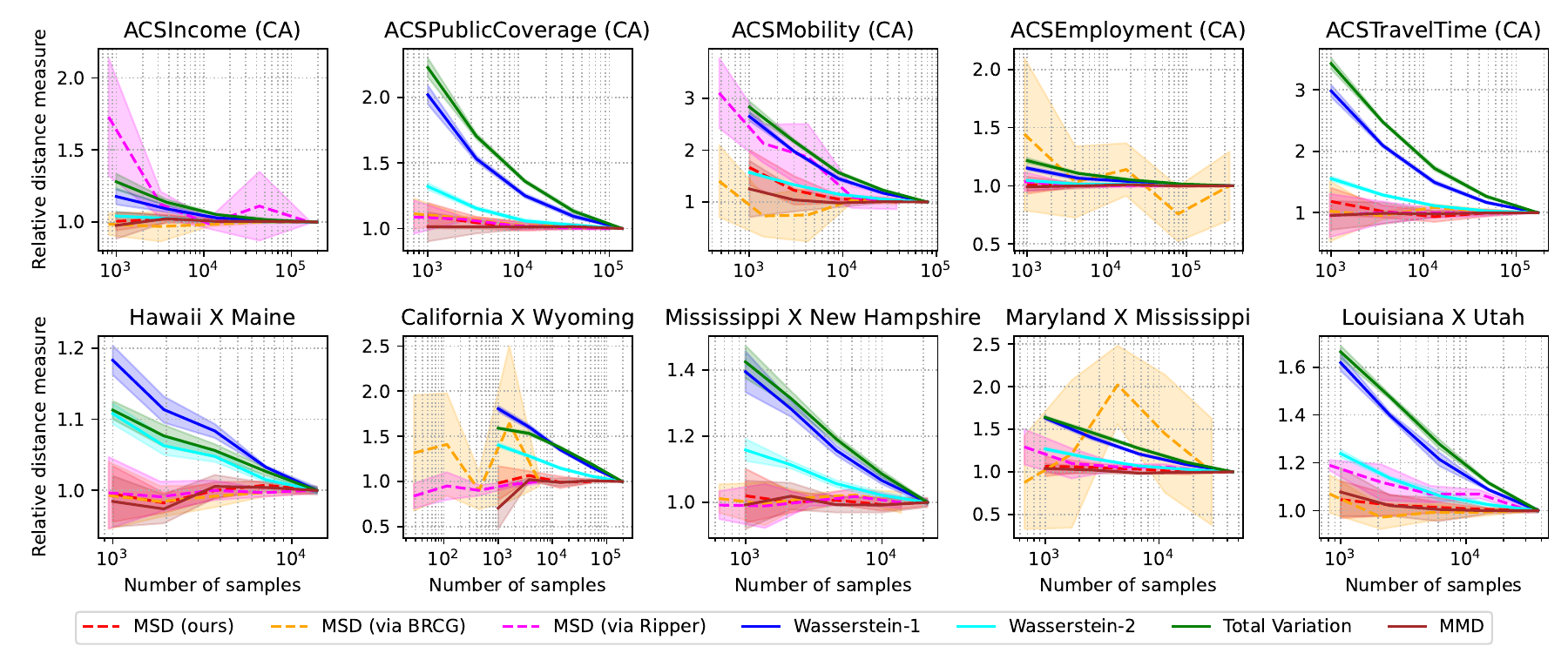}
    \caption{Mean relative distance measured by each method. The datasets in the top row are standard datasets provided by the folktables library. The bottom row shows datasets from the US Census, where we consider many protected attributes and compare their distributions on a pair of US states. Dashed lines represent methods that find a subgroup and for which we report the $\MSDc$ measure. The color bands represent the standard deviation. Ideally, each relative distance estimate should be 1, meaning that the estimate is the same as the estimate on the highest number of samples.}
    \label{fig:rel_dist}
    \Description{
    The figure contains 10 plots in 2 rows; on top are the convergence results for folktables datasets, and on the bottom are results for 5 state pair comparisons. The MIO-based MSD and MMD show good and fast convergence, BRCG and Ripper seem somewhat erratic, and Wasserstein metrics and Total Variation converge, but notably slower.
    }
\end{figure*}

The means over 5 random seeds are presented in Figure \ref{fig:rel_dist}. We report Ripper and BRCG results on different numbers of samples because these methods require equal sample sizes for both distributions, so we subsample the distribution for which we have more samples. This is especially pronounced in the California X Wyoming comparison due to the difference in population size. 
The BRCG and Ripper often vary in the quality of the distance estimate and have a rather high standard deviation. This is due to the methods being influenced by the randomness of the data or the algorithm. 
%
Clearly, TV and Wasserstein-based measures do not converge quickly to the final value. It is possible that the best estimate is still far from the true distance measure. This suggests that we cannot be certain of the true value of the distance measure. 

Finally, MMD shows comparable performance to our MIO-based $\MSDc$, both in deviation and in convergence. While this seems to empirically point to low sample complexity on these datasets, one cannot be certain that this holds always. In contrast, this is the case for our $\MSDc$, with high probability. 

The MIO-based $\MSDc$ almost always finds the global optimum within 10 minutes. 
The only exceptions were 13 (out of 250) seeded setups on three datasets with many protected attributes and samples, which could be expected.
Luckily, we show that we do not need a high number of samples to provide a good estimate of $\MSDc$. 
Additionally, providing the solver with an optimal solution on the smaller sample size could be used to prune some branches in the solver and would likely help with runtime, even on bigger instances.
In addition, our method's strong performance means that we also recover the correct subgroup with maximal distance, even with exponentially fewer samples than the maximum over the dataset. 

\subsubsection{Comparing algorithms for $\MSDc$ (Q3)}
\begin{figure*}[t]
    \centering
    \includegraphics[width=\linewidth]{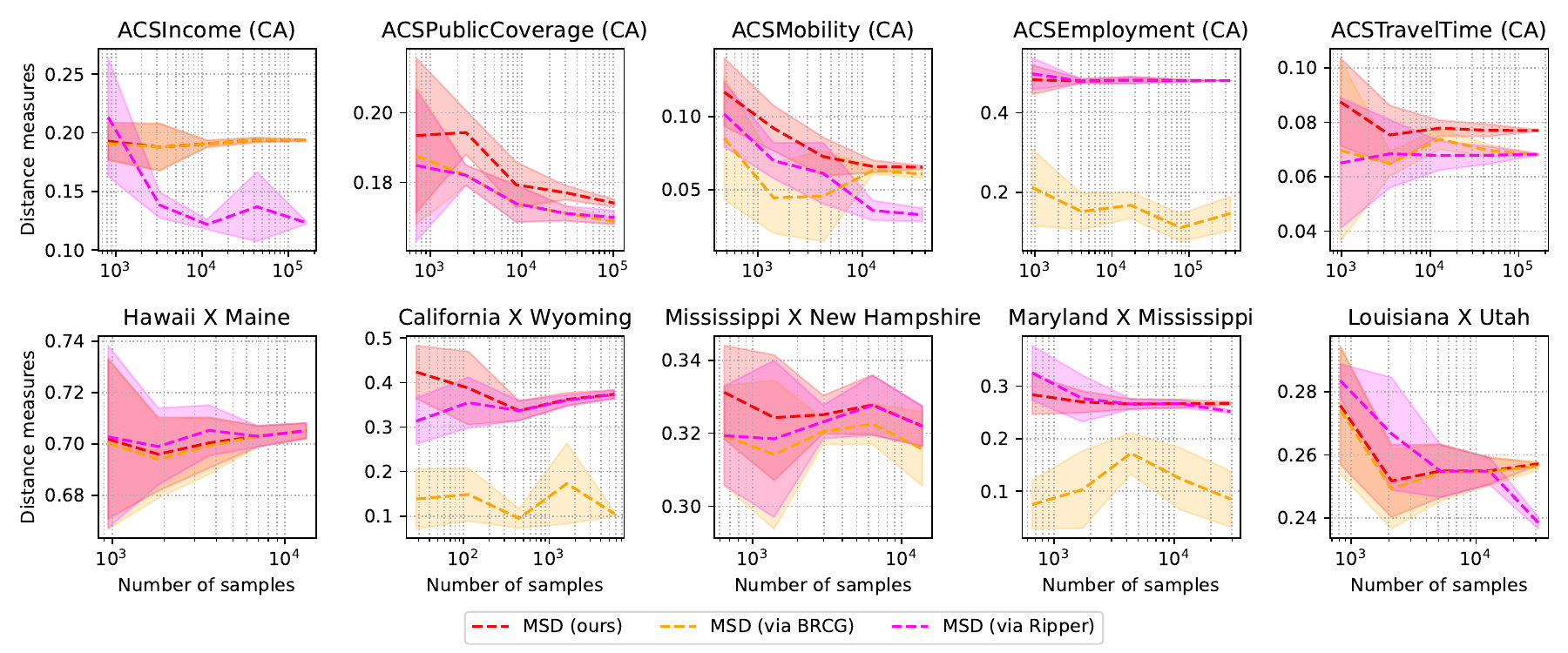}
    \caption{Evaluation of $\MSDc$ using three methods. The lines represent mean values, and the color bands represent the standard deviation.  MIO-based solutions (ours) dominate the solutions of non-globally optimal solvers. This is not the case for some runs with low sample sizes due to the fact that BRCG and Ripper consider subgroups with fewer than 10 samples.}
    \Description{
    The figure contains 10 plots in 2 rows; on top are the distance results for folktables datasets, and on the bottom are results for 5 state pair comparisons. The MIO MSD is mostly on top; BRCG and Ripper reach the same value in some plots, but in other plots, they differ by a lot.
    }
    \label{fig:MSDcomp}
\end{figure*}

As explained in Section \ref{sec:MSDclassification}, one can utilize standard DNF (or conjunction) learning algorithms to evaluate $\MSDc$.
We compare our proposed MIO formulation to Ripper and BRCG specifically, in Figure \ref{fig:MSDcomp}. We run the proposed method on the data with a balanced number of samples (by subsampling) of the two distributions to ensure comparability. The proposed method does not finish within 10 minutes for only three configurations. In all other cases, it returns global optima. 

The MIO-based $\MSDc$ value is always higher than (or equal to) the competitor methods due to the global optimality of the solutions. This does not always hold in some lower-sample configurations, which is due to the fact that BRCG and Ripper were not constrained to consider only subgroups with fewer than 10 samples.
More importantly, the $\MSDc$ values found by BRCG and Ripper are less stable with increasing sample size and often do not converge to the globally maximal value. This illustrates the unsuitability of BRCG and Ripper, since one does not obtain guarantees.  

Furthermore, this guarantee of global maximality enables two things. Firstly, if the MSD is below some given threshold, we can be sure that no subgroup violates the threshold, using the empirical distributions. This threshold can be computed by setting a probability level $\delta$ and utilizing Theorem \ref{thm:main_sample_thm} to obtain a guarantee with probability $\delta$. Secondly, we obtain the most disadvantaged subgroup. 
For example, the group with the highest $\MSDc$ for the task of estimating whether the commute to work takes more than 20 minutes (ACSTravelTime) was white women. 
All found groups are listed in Appendix \ref{app:subgroup_list}.

\section{Conclusion}
\label{sec:conclusion}
We pointed out the difficulties of evaluating the distance of probability distributions by current methods, especially in the context of evaluating bias for intersectional subgroups. 
We defined the Maximum Subgroup Discrepancy ($\MSDc$) and proved its linear sample complexity, in contrast to the exponential sample complexities of existing methods. We validated this on prediction tasks from real US Census data with varying numbers of protected attributes and sample sizes.
Only MMD shows empirically comparable sample complexity to our MIO-based $\MSDc$, but it does not provide as good an interpretation that could be useful in bias evaluation. 

The MIO formulation also outperforms more general DNF learners, which lack guarantees due to not being global optimizers.
In addition, the global optimality of MIO does not necessarily come at a disadvantage, as is otherwise common. That is because the linear sample complexity means that we do not need exponentially many samples, thus the formulation remains smaller and solves faster. 



\section*{Acknowledgment}
This work has received funding from the European Union’s Horizon Europe research and innovation
programme under grant agreement No. 101070568. 
Access to the computational infrastructure of the OP VVV-funded project CZ.02.1.01/0.0/0.0/16 019/0000765 “Research Center for Informatics” is also gratefully acknowledged. 
The first author also received funding from the Grant Agency of the Czech Technical University in Prague, grant No. SGS23/184/OHK3/3T/13.

\bibliographystyle{ACM-Reference-Format}
\bibliography{refs.bib}


\begin{thebibliography}{62}


\ifx \showCODEN    \undefined \def \showCODEN     #1{\unskip}     \fi
\ifx \showISBNx    \undefined \def \showISBNx     #1{\unskip}     \fi
\ifx \showISBNxiii \undefined \def \showISBNxiii  #1{\unskip}     \fi
\ifx \showISSN     \undefined \def \showISSN      #1{\unskip}     \fi
\ifx \showLCCN     \undefined \def \showLCCN      #1{\unskip}     \fi
\ifx \shownote     \undefined \def \shownote      #1{#1}          \fi
\ifx \showarticletitle \undefined \def \showarticletitle #1{#1}   \fi
\ifx \showURL      \undefined \def \showURL       {\relax}        \fi
\providecommand\bibfield[2]{#2}
\providecommand\bibinfo[2]{#2}
\providecommand\natexlab[1]{#1}
\providecommand\showeprint[2][]{arXiv:#2}

\bibitem[Ahn et~al\mbox{.}(2023)]%
        {ahn2023mitigating}
\bibfield{author}{\bibinfo{person}{Sumyeong Ahn}, \bibinfo{person}{Seongyoon Kim}, {and} \bibinfo{person}{Se-Young Yun}.} \bibinfo{year}{2023}\natexlab{}.
\newblock \showarticletitle{Mitigating Dataset Bias by Using Per-Sample Gradient}. In \bibinfo{booktitle}{\emph{The Eleventh International Conference on Learning Representations}}.
\newblock
\urldef\tempurl%
\url{https://openreview.net/forum?id=7mgUec-7GMv}
\showURL{%
\tempurl}


\bibitem[Arya et~al\mbox{.}(2019)]%
        {arya2019one}
\bibfield{author}{\bibinfo{person}{Vijay Arya}, \bibinfo{person}{Rachel K.~E. Bellamy}, \bibinfo{person}{Pin-Yu Chen}, \bibinfo{person}{Amit Dhurandhar}, \bibinfo{person}{Michael Hind}, \bibinfo{person}{Samuel~C. Hoffman}, \bibinfo{person}{Stephanie Houde}, \bibinfo{person}{Q.~Vera Liao}, \bibinfo{person}{Ronny Luss}, \bibinfo{person}{Aleksandra Mojsilovi\'c}, \bibinfo{person}{Sami Mourad}, \bibinfo{person}{Pablo Pedemonte}, \bibinfo{person}{Ramya Raghavendra}, \bibinfo{person}{John Richards}, \bibinfo{person}{Prasanna Sattigeri}, \bibinfo{person}{Karthikeyan Shanmugam}, \bibinfo{person}{Moninder Singh}, \bibinfo{person}{Kush~R. Varshney}, \bibinfo{person}{Dennis Wei}, {and} \bibinfo{person}{Yunfeng Zhang}.} \bibinfo{year}{2019}\natexlab{}.
\newblock \bibinfo{title}{One explanation does not fit all: A toolkit and taxonomy of ai explainability techniques}.
\newblock


\bibitem[Bahng et~al\mbox{.}(2020)]%
        {10.5555/3524938.3524988}
\bibfield{author}{\bibinfo{person}{Hyojin Bahng}, \bibinfo{person}{Sanghyuk Chun}, \bibinfo{person}{Sangdoo Yun}, \bibinfo{person}{Jaegul Choo}, {and} \bibinfo{person}{Seong~Joon Oh}.} \bibinfo{year}{2020}\natexlab{}.
\newblock \showarticletitle{Learning de-biased representations with biased representations}. In \bibinfo{booktitle}{\emph{Proceedings of the 37th International Conference on Machine Learning}} \emph{(\bibinfo{series}{ICML'20})}. \bibinfo{publisher}{JMLR.org}, Article \bibinfo{articleno}{50}, \bibinfo{numpages}{12}~pages.
\newblock


\bibitem[Bhattacharyya et~al\mbox{.}(2023)]%
        {bhattacharyya2022approximating}
\bibfield{author}{\bibinfo{person}{Arnab Bhattacharyya}, \bibinfo{person}{Sutanu Gayen}, \bibinfo{person}{Kuldeep~S. Meel}, \bibinfo{person}{Dimitrios Myrisiotis}, \bibinfo{person}{A. Pavan}, {and} \bibinfo{person}{N.~V. Vinodchandran}.} \bibinfo{year}{2023}\natexlab{}.
\newblock \showarticletitle{On approximating total variation distance}. In \bibinfo{booktitle}{\emph{Proceedings of the Thirty-Second International Joint Conference on Artificial Intelligence}} (Macao, P.R.China) \emph{(\bibinfo{series}{IJCAI '23})}. Article \bibinfo{articleno}{387}, \bibinfo{numpages}{9}~pages.
\newblock
\showISBNx{978-1-956792-03-4}
\href{https://doi.org/10.24963/ijcai.2023/387}{doi:\nolinkurl{10.24963/ijcai.2023/387}}


\bibitem[Bynum et~al\mbox{.}(2021)]%
        {bynumPyomoOptimizationModeling2021}
\bibfield{author}{\bibinfo{person}{Michael~L Bynum}, \bibinfo{person}{Gabriel~A Hackebeil}, \bibinfo{person}{William~E Hart}, \bibinfo{person}{Carl~D Laird}, \bibinfo{person}{Bethany~L Nicholson}, \bibinfo{person}{John~D Siirola}, \bibinfo{person}{Jean-Paul Watson}, \bibinfo{person}{David~L Woodruff}, {et~al\mbox{.}}} \bibinfo{year}{2021}\natexlab{}.
\newblock \bibinfo{booktitle}{\emph{Pyomo - {{Optimization Modeling}} in {{Python}}, 3rd {{Edition}}}}. Vol.~\bibinfo{volume}{67}.
\newblock \bibinfo{publisher}{Springer}.
\newblock


\bibitem[Chan et~al\mbox{.}(2014)]%
        {chan2014optimal}
\bibfield{author}{\bibinfo{person}{Siu-On Chan}, \bibinfo{person}{Ilias Diakonikolas}, \bibinfo{person}{Paul Valiant}, {and} \bibinfo{person}{Gregory Valiant}.} \bibinfo{year}{2014}\natexlab{}.
\newblock \showarticletitle{Optimal algorithms for testing closeness of discrete distributions}. In \bibinfo{booktitle}{\emph{Proceedings of the twenty-fifth annual ACM-SIAM symposium on Discrete algorithms}}. SIAM, \bibinfo{pages}{1193--1203}.
\newblock


\bibitem[Chen et~al\mbox{.}(2023)]%
        {chen2023sample}
\bibfield{author}{\bibinfo{person}{Ziyu Chen}, \bibinfo{person}{Markos Katsoulakis}, \bibinfo{person}{Luc Rey-Bellet}, {and} \bibinfo{person}{Wei Zhu}.} \bibinfo{year}{2023}\natexlab{}.
\newblock \showarticletitle{Sample complexity of probability divergences under group symmetry}. In \bibinfo{booktitle}{\emph{International Conference on Machine Learning}}. PMLR, \bibinfo{pages}{4713--4734}.
\newblock


\bibitem[Chen et~al\mbox{.}(2024)]%
        {10.1145/3597503.3639083}
\bibfield{author}{\bibinfo{person}{Zhenpeng Chen}, \bibinfo{person}{Jie~M. Zhang}, \bibinfo{person}{Federica Sarro}, {and} \bibinfo{person}{Mark Harman}.} \bibinfo{year}{2024}\natexlab{}.
\newblock \showarticletitle{Fairness Improvement with Multiple Protected Attributes: How Far Are We?}. In \bibinfo{booktitle}{\emph{Proceedings of the IEEE/ACM 46th International Conference on Software Engineering}} (Lisbon, Portugal) \emph{(\bibinfo{series}{ICSE '24})}. \bibinfo{publisher}{Association for Computing Machinery}, \bibinfo{address}{New York, NY, USA}, Article \bibinfo{articleno}{160}, \bibinfo{numpages}{13}~pages.
\newblock
\showISBNx{9798400702174}
\href{https://doi.org/10.1145/3597503.3639083}{doi:\nolinkurl{10.1145/3597503.3639083}}


\bibitem[Cohen(1995)]%
        {ripper}
\bibfield{author}{\bibinfo{person}{William~W. Cohen}.} \bibinfo{year}{1995}\natexlab{}.
\newblock \showarticletitle{Fast effective rule induction}. In \bibinfo{booktitle}{\emph{Proceedings of the Twelfth International Conference on International Conference on Machine Learning}} (Tahoe City, California, USA) \emph{(\bibinfo{series}{ICML'95})}. \bibinfo{publisher}{Morgan Kaufmann Publishers Inc.}, \bibinfo{address}{San Francisco, CA, USA}, \bibinfo{pages}{115–123}.
\newblock
\showISBNx{1558603778}


\bibitem[Collins and Bilge(2020)]%
        {collins2020intersectionality}
\bibfield{author}{\bibinfo{person}{Patricia~Hill Collins} {and} \bibinfo{person}{Sirma Bilge}.} \bibinfo{year}{2020}\natexlab{}.
\newblock \bibinfo{booktitle}{\emph{Intersectionality}}.
\newblock \bibinfo{publisher}{John Wiley \& Sons}.
\newblock


\bibitem[Crenshaw(2013)]%
        {crenshaw2013demarginalizing}
\bibfield{author}{\bibinfo{person}{Kimberl{\'e} Crenshaw}.} \bibinfo{year}{2013}\natexlab{}.
\newblock \showarticletitle{Demarginalizing the intersection of race and sex: A black feminist critique of antidiscrimination doctrine, feminist theory and antiracist politics}.
\newblock In \bibinfo{booktitle}{\emph{Feminist legal theories}}. \bibinfo{publisher}{Routledge}, \bibinfo{pages}{23--51}.
\newblock


\bibitem[Daniely and Shalev-Shwartz(2016)]%
        {daniely2016complexity}
\bibfield{author}{\bibinfo{person}{Amit Daniely} {and} \bibinfo{person}{Shai Shalev-Shwartz}.} \bibinfo{year}{2016}\natexlab{}.
\newblock \showarticletitle{Complexity theoretic limitations on learning dnf’s}. In \bibinfo{booktitle}{\emph{Conference on Learning Theory}}. PMLR, \bibinfo{pages}{815--830}.
\newblock


\bibitem[Dash et~al\mbox{.}(2018)]%
        {dash2018boolean}
\bibfield{author}{\bibinfo{person}{Sanjeeb Dash}, \bibinfo{person}{Oktay Gunluk}, {and} \bibinfo{person}{Dennis Wei}.} \bibinfo{year}{2018}\natexlab{}.
\newblock \showarticletitle{Boolean decision rules via column generation}.
\newblock \bibinfo{journal}{\emph{Advances in neural information processing systems}}  \bibinfo{volume}{31} (\bibinfo{year}{2018}).
\newblock


\bibitem[Devroye et~al\mbox{.}(2018)]%
        {devroye2018total}
\bibfield{author}{\bibinfo{person}{Luc Devroye}, \bibinfo{person}{Abbas Mehrabian}, {and} \bibinfo{person}{Tommy Reddad}.} \bibinfo{year}{2018}\natexlab{}.
\newblock \showarticletitle{The total variation distance between high-dimensional Gaussians with the same mean}.
\newblock \bibinfo{journal}{\emph{arXiv preprint arXiv:1810.08693}} (\bibinfo{year}{2018}).
\newblock


\bibitem[Ding et~al\mbox{.}(2021)]%
        {ding2021retiring}
\bibfield{author}{\bibinfo{person}{Frances Ding}, \bibinfo{person}{Moritz Hardt}, \bibinfo{person}{John Miller}, {and} \bibinfo{person}{Ludwig Schmidt}.} \bibinfo{year}{2021}\natexlab{}.
\newblock \showarticletitle{Retiring Adult: New Datasets for Fair Machine Learning}.
\newblock \bibinfo{journal}{\emph{Advances in Neural Information Processing Systems}}  \bibinfo{volume}{34} (\bibinfo{year}{2021}).
\newblock


\bibitem[Dominguez-Catena et~al\mbox{.}(2024)]%
        {10420507}
\bibfield{author}{\bibinfo{person}{Iris Dominguez-Catena}, \bibinfo{person}{Daniel Paternain}, {and} \bibinfo{person}{Mikel Galar}.} \bibinfo{year}{2024}\natexlab{}.
\newblock \showarticletitle{Metrics for Dataset Demographic Bias: A Case Study on Facial Expression Recognition}.
\newblock \bibinfo{journal}{\emph{IEEE Transactions on Pattern Analysis and Machine Intelligence}} \bibinfo{volume}{46}, \bibinfo{number}{8} (\bibinfo{year}{2024}), \bibinfo{pages}{5209--5226}.
\newblock
\href{https://doi.org/10.1109/TPAMI.2024.3361979}{doi:\nolinkurl{10.1109/TPAMI.2024.3361979}}


\bibitem[Dudley(1969)]%
        {dudley1969speed}
\bibfield{author}{\bibinfo{person}{Richard~Mansfield Dudley}.} \bibinfo{year}{1969}\natexlab{}.
\newblock \showarticletitle{The speed of mean Glivenko-Cantelli convergence}.
\newblock \bibinfo{journal}{\emph{The Annals of Mathematical Statistics}} \bibinfo{volume}{40}, \bibinfo{number}{1} (\bibinfo{year}{1969}), \bibinfo{pages}{40--50}.
\newblock


\bibitem[Feng et~al\mbox{.}(2023)]%
        {feng2023simple}
\bibfield{author}{\bibinfo{person}{Weiming Feng}, \bibinfo{person}{Heng Guo}, \bibinfo{person}{Mark Jerrum}, {and} \bibinfo{person}{Jiaheng Wang}.} \bibinfo{year}{2023}\natexlab{}.
\newblock \showarticletitle{A simple polynomial-time approximation algorithm for the total variation distance between two product distributions}.
\newblock \bibinfo{journal}{\emph{TheoretiCS}}  \bibinfo{volume}{2} (\bibinfo{year}{2023}).
\newblock


\bibitem[Flamary et~al\mbox{.}(2021)]%
        {flamary2021pot}
\bibfield{author}{\bibinfo{person}{R{\'e}mi Flamary}, \bibinfo{person}{Nicolas Courty}, \bibinfo{person}{Alexandre Gramfort}, \bibinfo{person}{Mokhtar~Z. Alaya}, \bibinfo{person}{Aur{\'e}lie Boisbunon}, \bibinfo{person}{Stanislas Chambon}, \bibinfo{person}{Laetitia Chapel}, \bibinfo{person}{Adrien Corenflos}, \bibinfo{person}{Kilian Fatras}, \bibinfo{person}{Nemo Fournier}, \bibinfo{person}{L{\'e}o Gautheron}, \bibinfo{person}{Nathalie~T.H. Gayraud}, \bibinfo{person}{Hicham Janati}, \bibinfo{person}{Alain Rakotomamonjy}, \bibinfo{person}{Ievgen Redko}, \bibinfo{person}{Antoine Rolet}, \bibinfo{person}{Antony Schutz}, \bibinfo{person}{Vivien Seguy}, \bibinfo{person}{Danica~J. Sutherland}, \bibinfo{person}{Romain Tavenard}, \bibinfo{person}{Alexander Tong}, {and} \bibinfo{person}{Titouan Vayer}.} \bibinfo{year}{2021}\natexlab{}.
\newblock \showarticletitle{POT: Python Optimal Transport}.
\newblock \bibinfo{journal}{\emph{Journal of Machine Learning Research}} \bibinfo{volume}{22}, \bibinfo{number}{78} (\bibinfo{year}{2021}), \bibinfo{pages}{1--8}.
\newblock
\urldef\tempurl%
\url{http://jmlr.org/papers/v22/20-451.html}
\showURL{%
\tempurl}


\bibitem[Foulds et~al\mbox{.}(2020)]%
        {foulds2020intersectional}
\bibfield{author}{\bibinfo{person}{James~R Foulds}, \bibinfo{person}{Rashidul Islam}, \bibinfo{person}{Kamrun~Naher Keya}, {and} \bibinfo{person}{Shimei Pan}.} \bibinfo{year}{2020}\natexlab{}.
\newblock \showarticletitle{An intersectional definition of fairness}. In \bibinfo{booktitle}{\emph{2020 IEEE 36th International Conference on Data Engineering (ICDE)}}. IEEE, \bibinfo{pages}{1918--1921}.
\newblock


\bibitem[Fournier and Guillin(2015)]%
        {fournier2015rate}
\bibfield{author}{\bibinfo{person}{Nicolas Fournier} {and} \bibinfo{person}{Arnaud Guillin}.} \bibinfo{year}{2015}\natexlab{}.
\newblock \showarticletitle{On the rate of convergence in Wasserstein distance of the empirical measure}.
\newblock \bibinfo{journal}{\emph{Probability theory and related fields}} \bibinfo{volume}{162}, \bibinfo{number}{3} (\bibinfo{year}{2015}), \bibinfo{pages}{707--738}.
\newblock


\bibitem[Gohar and Cheng(2023)]%
        {gohar2023survey}
\bibfield{author}{\bibinfo{person}{Usman Gohar} {and} \bibinfo{person}{Lu Cheng}.} \bibinfo{year}{2023}\natexlab{}.
\newblock \showarticletitle{A Survey on Intersectional Fairness in Machine Learning: Notions, Mitigation, and Challenges}. In \bibinfo{booktitle}{\emph{Proceedings of the Thirty-Second International Joint Conference on Artificial Intelligence, {IJCAI-23}}}, \bibfield{editor}{\bibinfo{person}{Edith Elkind}} (Ed.). \bibinfo{publisher}{International Joint Conferences on Artificial Intelligence Organization}, \bibinfo{pages}{6619--6627}.
\newblock
\href{https://doi.org/10.24963/ijcai.2023/742}{doi:\nolinkurl{10.24963/ijcai.2023/742}}
\newblock
\shownote{Survey Track}.


\bibitem[Gretton et~al\mbox{.}(2012)]%
        {gretton2012kernel}
\bibfield{author}{\bibinfo{person}{Arthur Gretton}, \bibinfo{person}{Karsten~M Borgwardt}, \bibinfo{person}{Malte~J Rasch}, \bibinfo{person}{Bernhard Sch{\"o}lkopf}, {and} \bibinfo{person}{Alexander Smola}.} \bibinfo{year}{2012}\natexlab{}.
\newblock \showarticletitle{A kernel two-sample test}.
\newblock \bibinfo{journal}{\emph{The Journal of Machine Learning Research}} \bibinfo{volume}{13}, \bibinfo{number}{1} (\bibinfo{year}{2012}), \bibinfo{pages}{723--773}.
\newblock


\bibitem[{Gurobi Optimization, LLC}(2023)]%
        {gurobi}
\bibfield{author}{\bibinfo{person}{{Gurobi Optimization, LLC}}.} \bibinfo{year}{2023}\natexlab{}.
\newblock \bibinfo{title}{Gurobi Optimizer Reference Manual}.
\newblock
\urldef\tempurl%
\url{https://www.gurobi.com}
\showURL{%
\tempurl}


\bibitem[Haldar et~al\mbox{.}(2023)]%
        {haldar2023interpretable}
\bibfield{author}{\bibinfo{person}{Swagatam Haldar}, \bibinfo{person}{Diptikalyan Saha}, \bibinfo{person}{Dennis Wei}, \bibinfo{person}{Rahul Nair}, {and} \bibinfo{person}{Elizabeth~M Daly}.} \bibinfo{year}{2023}\natexlab{}.
\newblock \showarticletitle{Interpretable differencing of machine learning models}. In \bibinfo{booktitle}{\emph{Uncertainty in Artificial Intelligence}}. PMLR, \bibinfo{pages}{788--797}.
\newblock


\bibitem[Hellinger(1909)]%
        {hellinger1909neue}
\bibfield{author}{\bibinfo{person}{Ernst Hellinger}.} \bibinfo{year}{1909}\natexlab{}.
\newblock \showarticletitle{Neue begr{\"u}ndung der theorie quadratischer formen von unendlichvielen ver{\"a}nderlichen.}
\newblock \bibinfo{journal}{\emph{Journal f{\"u}r die reine und angewandte Mathematik}} \bibinfo{volume}{1909}, \bibinfo{number}{136} (\bibinfo{year}{1909}), \bibinfo{pages}{210--271}.
\newblock


\bibitem[Hsu et~al\mbox{.}(2024)]%
        {hsu2024polynomial}
\bibfield{author}{\bibinfo{person}{Daniel Hsu}, \bibinfo{person}{Jizhou Huang}, {and} \bibinfo{person}{Brendan Juba}.} \bibinfo{year}{2024}\natexlab{}.
\newblock \showarticletitle{Polynomial time auditing of statistical subgroup fairness for Gaussian data}.
\newblock \bibinfo{journal}{\emph{CoRR}}  \bibinfo{volume}{abs/2401.16439} (\bibinfo{year}{2024}).
\newblock
\urldef\tempurl%
\url{https://doi.org/10.48550/arXiv.2401.16439}
\showURL{%
\tempurl}


\bibitem[{IEEE Standards Association}(2025)]%
        {10851955}
\bibfield{author}{\bibinfo{person}{{IEEE Standards Association}}.} \bibinfo{year}{2025}\natexlab{}.
\newblock \showarticletitle{IEEE Standard for Algorithmic Bias Considerations}.
\newblock \bibinfo{journal}{\emph{IEEE Std 7003-2024}} (\bibinfo{year}{2025}), \bibinfo{pages}{1--59}.
\newblock
\href{https://doi.org/10.1109/IEEESTD.2025.10851955}{doi:\nolinkurl{10.1109/IEEESTD.2025.10851955}}


\bibitem[Jordan(1881)]%
        {zbMATH02706589}
\bibfield{author}{\bibinfo{person}{C. Jordan}.} \bibinfo{year}{1881}\natexlab{}.
\newblock \showarticletitle{On {Fourier} series}.
\newblock \bibinfo{journal}{\emph{C. R. Acad. Sci., Paris}}  \bibinfo{volume}{92} (\bibinfo{year}{1881}), \bibinfo{pages}{228--230}.
\newblock
\showISSN{0001-4036}


\bibitem[Kandiros et~al\mbox{.}(2023)]%
        {kandiros2023learning}
\bibfield{author}{\bibinfo{person}{Vardis Kandiros}, \bibinfo{person}{Constantinos Daskalakis}, \bibinfo{person}{Yuval Dagan}, {and} \bibinfo{person}{Davin Choo}.} \bibinfo{year}{2023}\natexlab{}.
\newblock \showarticletitle{Learning and Testing Latent-Tree Ising Models Efficiently}. In \bibinfo{booktitle}{\emph{The Thirty Sixth Annual Conference on Learning Theory}}. PMLR, \bibinfo{pages}{1666--1729}.
\newblock


\bibitem[Kearns et~al\mbox{.}(2018)]%
        {kearns2018preventing}
\bibfield{author}{\bibinfo{person}{Michael Kearns}, \bibinfo{person}{Seth Neel}, \bibinfo{person}{Aaron Roth}, {and} \bibinfo{person}{Zhiwei~Steven Wu}.} \bibinfo{year}{2018}\natexlab{}.
\newblock \showarticletitle{Preventing fairness gerrymandering: Auditing and learning for subgroup fairness}. In \bibinfo{booktitle}{\emph{International conference on machine learning}}. PMLR, \bibinfo{pages}{2564--2572}.
\newblock


\bibitem[Kearns et~al\mbox{.}(2019)]%
        {kearns2019empirical}
\bibfield{author}{\bibinfo{person}{Michael Kearns}, \bibinfo{person}{Seth Neel}, \bibinfo{person}{Aaron Roth}, {and} \bibinfo{person}{Zhiwei~Steven Wu}.} \bibinfo{year}{2019}\natexlab{}.
\newblock \showarticletitle{An empirical study of rich subgroup fairness for machine learning}. In \bibinfo{booktitle}{\emph{Proceedings of the conference on fairness, accountability, and transparency}}. \bibinfo{pages}{100--109}.
\newblock


\bibitem[Klivans and Servedio(2001)]%
        {klivans2001learning}
\bibfield{author}{\bibinfo{person}{Adam~R Klivans} {and} \bibinfo{person}{Rocco Servedio}.} \bibinfo{year}{2001}\natexlab{}.
\newblock \showarticletitle{Learning DNF in time}. In \bibinfo{booktitle}{\emph{Proceedings of the thirty-third annual ACM symposium on Theory of computing}}. \bibinfo{pages}{258--265}.
\newblock


\bibitem[Koch et~al\mbox{.}(2022)]%
        {kochProgressMathematicalProgramming2022}
\bibfield{author}{\bibinfo{person}{Thorsten Koch}, \bibinfo{person}{Timo Berthold}, \bibinfo{person}{Jaap Pedersen}, {and} \bibinfo{person}{Charlie Vanaret}.} \bibinfo{year}{2022}\natexlab{}.
\newblock \showarticletitle{Progress in Mathematical Programming Solvers from 2001 to 2020}.
\newblock \bibinfo{journal}{\emph{EURO Journal on Computational Optimization}}  \bibinfo{volume}{10} (\bibinfo{date}{Jan.} \bibinfo{year}{2022}), \bibinfo{pages}{100031}.
\newblock
\showISSN{2192-4406}
\href{https://doi.org/10.1016/j.ejco.2022.100031}{doi:\nolinkurl{10.1016/j.ejco.2022.100031}}


\bibitem[Kullback and Leibler(1951)]%
        {KL}
\bibfield{author}{\bibinfo{person}{S. Kullback} {and} \bibinfo{person}{R.~A. Leibler}.} \bibinfo{year}{1951}\natexlab{}.
\newblock \showarticletitle{{On Information and Sufficiency}}.
\newblock \bibinfo{journal}{\emph{The Annals of Mathematical Statistics}} \bibinfo{volume}{22}, \bibinfo{number}{1} (\bibinfo{year}{1951}), \bibinfo{pages}{79 -- 86}.
\newblock
\href{https://doi.org/10.1214/aoms/1177729694}{doi:\nolinkurl{10.1214/aoms/1177729694}}


\bibitem[Lawless et~al\mbox{.}(2023)]%
        {lawlessInterpretableFairBoolean2023}
\bibfield{author}{\bibinfo{person}{Connor Lawless}, \bibinfo{person}{Sanjeeb Dash}, \bibinfo{person}{Oktay Gunluk}, {and} \bibinfo{person}{Dennis Wei}.} \bibinfo{year}{2023}\natexlab{}.
\newblock \showarticletitle{Interpretable and {{Fair Boolean Rule Sets}} via {{Column Generation}}}.
\newblock \bibinfo{journal}{\emph{Journal of Machine Learning Research}} \bibinfo{volume}{24}, \bibinfo{number}{229} (\bibinfo{year}{2023}), \bibinfo{pages}{1--50}.
\newblock
\showISSN{1533-7928}


\bibitem[Lee et~al\mbox{.}(2024)]%
        {lee2023computability}
\bibfield{author}{\bibinfo{person}{Yunseok Lee}, \bibinfo{person}{Holger Boche}, {and} \bibinfo{person}{Gitta Kutyniok}.} \bibinfo{year}{2024}\natexlab{}.
\newblock \showarticletitle{Computability of Optimizers}.
\newblock \bibinfo{journal}{\emph{IEEE Transactions on Information Theory}} \bibinfo{volume}{70}, \bibinfo{number}{4} (\bibinfo{year}{2024}), \bibinfo{pages}{2967--2983}.
\newblock
\href{https://doi.org/10.1109/TIT.2023.3347071}{doi:\nolinkurl{10.1109/TIT.2023.3347071}}


\bibitem[Liu et~al\mbox{.}(2019)]%
        {liu2018rate}
\bibfield{author}{\bibinfo{person}{Anning Liu}, \bibinfo{person}{Jian-Guo Liu}, {and} \bibinfo{person}{Yulong Lu}.} \bibinfo{year}{2019}\natexlab{}.
\newblock \showarticletitle{On the rate of convergence of empirical measure in $\infty$-wasserstein distance for unbounded density function}.
\newblock \bibinfo{journal}{\emph{Quart. Appl. Math.}} \bibinfo{volume}{77}, \bibinfo{number}{4} (\bibinfo{year}{2019}), \bibinfo{pages}{pp. 811--829}.
\newblock
\showISSN{0033569X, 15524485}
\urldef\tempurl%
\url{https://www.jstor.org/stable/26839553}
\showURL{%
\tempurl}


\bibitem[Malioutov and Varshney(2013)]%
        {malioutovExactRuleLearning2013}
\bibfield{author}{\bibinfo{person}{Dmitry Malioutov} {and} \bibinfo{person}{Kush Varshney}.} \bibinfo{year}{2013}\natexlab{}.
\newblock \showarticletitle{Exact {{Rule Learning}} via {{Boolean Compressed Sensing}}}. In \bibinfo{booktitle}{\emph{Proceedings of the 30th {{International Conference}} on {{Machine Learning}}}}. \bibinfo{publisher}{PMLR}, \bibinfo{pages}{765--773}.
\newblock
\showISSN{1938-7228}


\bibitem[Minsky and Papert(1988)]%
        {minsky1988perceptrons}
\bibfield{author}{\bibinfo{person}{Marvin~L Minsky} {and} \bibinfo{person}{Seymour~A Papert}.} \bibinfo{year}{1988}\natexlab{}.
\newblock \bibinfo{title}{Perceptrons: expanded edition}.
\newblock


\bibitem[Mohri et~al\mbox{.}(2018)]%
        {mohri2018foundations}
\bibfield{author}{\bibinfo{person}{Mehryar Mohri}, \bibinfo{person}{Afshin Rostamizadeh}, {and} \bibinfo{person}{Ameet Talwalkar}.} \bibinfo{year}{2018}\natexlab{}.
\newblock \bibinfo{booktitle}{\emph{Foundations of machine learning}}.
\newblock \bibinfo{publisher}{MIT press}.
\newblock


\bibitem[Nair et~al\mbox{.}(2021)]%
        {nair2021changed}
\bibfield{author}{\bibinfo{person}{Rahul Nair}, \bibinfo{person}{Massimiliano Mattetti}, \bibinfo{person}{Elizabeth Daly}, \bibinfo{person}{Dennis Wei}, \bibinfo{person}{Oznur Alkan}, {and} \bibinfo{person}{Yunfeng Zhang}.} \bibinfo{year}{2021}\natexlab{}.
\newblock \showarticletitle{What Changed? Interpretable Model Comparison.}. In \bibinfo{booktitle}{\emph{IJCAI}}. \bibinfo{pages}{2855--2861}.
\newblock


\bibitem[Nam et~al\mbox{.}(2020)]%
        {NEURIPS2020_eddc3427}
\bibfield{author}{\bibinfo{person}{Junhyun Nam}, \bibinfo{person}{Hyuntak Cha}, \bibinfo{person}{Sungsoo Ahn}, \bibinfo{person}{Jaeho Lee}, {and} \bibinfo{person}{Jinwoo Shin}.} \bibinfo{year}{2020}\natexlab{}.
\newblock \showarticletitle{Learning from Failure: De-biasing Classifier from Biased Classifier}. In \bibinfo{booktitle}{\emph{Advances in Neural Information Processing Systems}}, \bibfield{editor}{\bibinfo{person}{H.~Larochelle}, \bibinfo{person}{M.~Ranzato}, \bibinfo{person}{R.~Hadsell}, \bibinfo{person}{M.F. Balcan}, {and} \bibinfo{person}{H.~Lin}} (Eds.), Vol.~\bibinfo{volume}{33}. \bibinfo{publisher}{Curran Associates, Inc.}, \bibinfo{pages}{20673--20684}.
\newblock
\urldef\tempurl%
\url{https://proceedings.neurips.cc/paper_files/paper/2020/file/eddc3427c5d77843c2253f1e799fe933-Paper.pdf}
\showURL{%
\tempurl}


\bibitem[Neill et~al\mbox{.}(2013)]%
        {neil2013fast}
\bibfield{author}{\bibinfo{person}{Daniel Neill}, \bibinfo{person}{Edward McFowland}, {and} \bibinfo{person}{Huanian Zheng}.} \bibinfo{year}{2013}\natexlab{}.
\newblock \showarticletitle{Fast subset scan for multivariate event detection}.
\newblock \bibinfo{journal}{\emph{Statistics in medicine}}  \bibinfo{volume}{32} (\bibinfo{date}{06} \bibinfo{year}{2013}).
\newblock
\href{https://doi.org/10.1002/sim.5675}{doi:\nolinkurl{10.1002/sim.5675}}


\bibitem[Nguyen and Ho(2022)]%
        {NEURIPS2022_f02f1185}
\bibfield{author}{\bibinfo{person}{Khai Nguyen} {and} \bibinfo{person}{Nhat Ho}.} \bibinfo{year}{2022}\natexlab{}.
\newblock \showarticletitle{Amortized Projection Optimization for Sliced Wasserstein Generative Models}. In \bibinfo{booktitle}{\emph{Advances in Neural Information Processing Systems}}, \bibfield{editor}{\bibinfo{person}{S.~Koyejo}, \bibinfo{person}{S.~Mohamed}, \bibinfo{person}{A.~Agarwal}, \bibinfo{person}{D.~Belgrave}, \bibinfo{person}{K.~Cho}, {and} \bibinfo{person}{A.~Oh}} (Eds.), Vol.~\bibinfo{volume}{35}. \bibinfo{publisher}{Curran Associates, Inc.}, \bibinfo{pages}{36985--36998}.
\newblock
\urldef\tempurl%
\url{https://proceedings.neurips.cc/paper_files/paper/2022/file/f02f1185b97518ab5bd7ebde466992d3-Paper-Conference.pdf}
\showURL{%
\tempurl}


\bibitem[Panaretos and Zemel(2019)]%
        {panaretos2019statistical}
\bibfield{author}{\bibinfo{person}{Victor~M Panaretos} {and} \bibinfo{person}{Yoav Zemel}.} \bibinfo{year}{2019}\natexlab{}.
\newblock \showarticletitle{Statistical aspects of Wasserstein distances}.
\newblock \bibinfo{journal}{\emph{Annual review of statistics and its application}} \bibinfo{volume}{6}, \bibinfo{number}{1} (\bibinfo{year}{2019}), \bibinfo{pages}{405--431}.
\newblock


\bibitem[Parliament and Council(2024)]%
        {AIAct}
\bibfield{author}{\bibinfo{person}{European Parliament} {and} \bibinfo{person}{European Council}.} \bibinfo{year}{2024}\natexlab{}.
\newblock \showarticletitle{Regulation (EU) 2024/1689 of the {E}uropean {P}arliament and of the {C}ouncil of 13 {J}une 2024 laying down harmonised rules on artificial intelligence and amending {Regulations} (EC) No 300/2008, (EU) No 167/2013, (EU) No 168/2013, (EU) 2018/858, (EU) 2018/1139 and (EU) 2019/2144 and Directives 2014/90/EU, (EU) 2016/797 and (EU) 2020/1828 ({A}rtificial {I}ntelligence {A}ct)}.
\newblock \bibinfo{journal}{\emph{Official Journal}} \bibinfo{volume}{2024}, \bibinfo{number}{1689} (\bibinfo{year}{2024}), \bibinfo{pages}{1--144}.
\newblock
\urldef\tempurl%
\url{http://data.europa.eu/eli/reg/2024/1689/oj}
\showURL{%
\tempurl}


\bibitem[Schwartz et~al\mbox{.}(2022)]%
        {schwartz2022towards}
\bibfield{author}{\bibinfo{person}{Reva Schwartz}, \bibinfo{person}{Apostol Vassilev}, \bibinfo{person}{Kristen Greene}, \bibinfo{person}{Lori Perine}, \bibinfo{person}{Andrew Burt}, {and} \bibinfo{person}{Patrick Hall}.} \bibinfo{year}{2022}\natexlab{}.
\newblock \bibinfo{booktitle}{\emph{Towards a standard for identifying and managing bias in artificial intelligence}}. Vol.~\bibinfo{volume}{3}.
\newblock \bibinfo{publisher}{US Department of Commerce, National Institute of Standards and Technology}.
\newblock


\bibitem[Shalev-Shwartz and Ben-David(2014)]%
        {shalev2014understanding}
\bibfield{author}{\bibinfo{person}{Shai Shalev-Shwartz} {and} \bibinfo{person}{Shai Ben-David}.} \bibinfo{year}{2014}\natexlab{}.
\newblock \bibinfo{booktitle}{\emph{Understanding machine learning: From theory to algorithms}}.
\newblock \bibinfo{publisher}{Cambridge university press}.
\newblock


\bibitem[Speakman et~al\mbox{.}(2015)]%
        {speakman2015penalized}
\bibfield{author}{\bibinfo{person}{Skyler Speakman}, \bibinfo{person}{Sriram Somanchi}, \bibinfo{person}{Edward McFowland}, {and} \bibinfo{person}{Daniel Neill}.} \bibinfo{year}{2015}\natexlab{}.
\newblock \showarticletitle{Penalized Fast Subset Scanning}.
\newblock \bibinfo{journal}{\emph{Journal of Computational and Graphical Statistics}}  \bibinfo{volume}{25} (\bibinfo{date}{04} \bibinfo{year}{2015}), \bibinfo{pages}{00--00}.
\newblock
\href{https://doi.org/10.1080/10618600.2015.1029578}{doi:\nolinkurl{10.1080/10618600.2015.1029578}}


\bibitem[Su et~al\mbox{.}(2016)]%
        {suLearningSparseTwolevel2016}
\bibfield{author}{\bibinfo{person}{Guolong Su}, \bibinfo{person}{Dennis Wei}, \bibinfo{person}{Kush~R. Varshney}, {and} \bibinfo{person}{Dmitry~M. Malioutov}.} \bibinfo{year}{2016}\natexlab{}.
\newblock \showarticletitle{Learning Sparse Two-Level Boolean Rules}. In \bibinfo{booktitle}{\emph{2016 {{IEEE}} 26th {{International Workshop}} on {{Machine Learning}} for {{Signal Processing}} ({{MLSP}})}}. \bibinfo{pages}{1--6}.
\newblock
\href{https://doi.org/10.1109/MLSP.2016.7738856}{doi:\nolinkurl{10.1109/MLSP.2016.7738856}}


\bibitem[Tahmasebi and Jegelka(2023)]%
        {tahmasebisample}
\bibfield{author}{\bibinfo{person}{Behrooz Tahmasebi} {and} \bibinfo{person}{Stefanie Jegelka}.} \bibinfo{year}{2023}\natexlab{}.
\newblock \showarticletitle{Sample Complexity Bounds for Estimating the Wasserstein Distance under Invariances}.
\newblock \bibinfo{journal}{\emph{2nd Annual Workshop on Topology, Algebra, and Geometry in Machine Learning ({TAG-ML})}} (\bibinfo{year}{2023}).
\newblock
\urldef\tempurl%
\url{https://openreview.net/forum?id=3fpo7JBC27}
\showURL{%
\tempurl}


\bibitem[Tolstikhin et~al\mbox{.}(2016)]%
        {NIPS2016_5055cbf4}
\bibfield{author}{\bibinfo{person}{Ilya~O Tolstikhin}, \bibinfo{person}{Bharath~K. Sriperumbudur}, {and} \bibinfo{person}{Bernhard Sch\"{o}lkopf}.} \bibinfo{year}{2016}\natexlab{}.
\newblock \showarticletitle{Minimax Estimation of Maximum Mean Discrepancy with Radial Kernels}. In \bibinfo{booktitle}{\emph{Advances in Neural Information Processing Systems}}, \bibfield{editor}{\bibinfo{person}{D.~Lee}, \bibinfo{person}{M.~Sugiyama}, \bibinfo{person}{U.~Luxburg}, \bibinfo{person}{I.~Guyon}, {and} \bibinfo{person}{R.~Garnett}} (Eds.), Vol.~\bibinfo{volume}{29}. \bibinfo{publisher}{Curran Associates, Inc.}
\newblock
\urldef\tempurl%
\url{https://proceedings.neurips.cc/paper_files/paper/2016/file/5055cbf43fac3f7e2336b27310f0b9ef-Paper.pdf}
\showURL{%
\tempurl}


\bibitem[Tsybakov(2009)]%
        {tsybakov2009nonparametric}
\bibfield{author}{\bibinfo{person}{Alexandre~B Tsybakov}.} \bibinfo{year}{2009}\natexlab{}.
\newblock \showarticletitle{Nonparametric estimators}.
\newblock \bibinfo{journal}{\emph{Introduction to Nonparametric Estimation}} (\bibinfo{year}{2009}), \bibinfo{pages}{1--76}.
\newblock


\bibitem[Valiant and Valiant(2011)]%
        {valiant2011estimating}
\bibfield{author}{\bibinfo{person}{Gregory Valiant} {and} \bibinfo{person}{Paul Valiant}.} \bibinfo{year}{2011}\natexlab{}.
\newblock \showarticletitle{Estimating the unseen: an n/log (n)-sample estimator for entropy and support size, shown optimal via new CLTs}. In \bibinfo{booktitle}{\emph{Proceedings of the forty-third annual ACM symposium on Theory of computing}}. \bibinfo{pages}{685--694}.
\newblock


\bibitem[Vaserstein(1969)]%
        {vaserstein1969markov}
\bibfield{author}{\bibinfo{person}{Leonid~Nisonovich Vaserstein}.} \bibinfo{year}{1969}\natexlab{}.
\newblock \showarticletitle{Markov processes over denumerable products of spaces, describing large systems of automata}.
\newblock \bibinfo{journal}{\emph{Problemy Peredachi Informatsii}} \bibinfo{volume}{5}, \bibinfo{number}{3} (\bibinfo{year}{1969}), \bibinfo{pages}{64--72}.
\newblock


\bibitem[Wang and Rudin(2015)]%
        {wangLearningOptimizedOrs2015}
\bibfield{author}{\bibinfo{person}{Tong Wang} {and} \bibinfo{person}{Cynthia Rudin}.} \bibinfo{year}{2015}\natexlab{}.
\newblock \bibinfo{title}{Learning {{Optimized Or}}'s of {{And}}'s}.
\newblock
\href{https://doi.org/10.48550/arXiv.1511.02210}{doi:\nolinkurl{10.48550/arXiv.1511.02210}}
\showeprint[arxiv]{1511.02210}~[cs]


\bibitem[Weed and Bach(2019)]%
        {weed2019sharp}
\bibfield{author}{\bibinfo{person}{Jonathan Weed} {and} \bibinfo{person}{Francis Bach}.} \bibinfo{year}{2019}\natexlab{}.
\newblock \showarticletitle{{Sharp asymptotic and finite-sample rates of convergence of empirical measures in Wasserstein distance}}.
\newblock \bibinfo{journal}{\emph{Bernoulli}} \bibinfo{volume}{25}, \bibinfo{number}{4A} (\bibinfo{year}{2019}), \bibinfo{pages}{2620 -- 2648}.
\newblock
\href{https://doi.org/10.3150/18-BEJ1065}{doi:\nolinkurl{10.3150/18-BEJ1065}}


\bibitem[Wei et~al\mbox{.}(2019)]%
        {wei2019generalized}
\bibfield{author}{\bibinfo{person}{Dennis Wei}, \bibinfo{person}{Sanjeeb Dash}, \bibinfo{person}{Tian Gao}, {and} \bibinfo{person}{Oktay Gunluk}.} \bibinfo{year}{2019}\natexlab{}.
\newblock \showarticletitle{Generalized linear rule models}. In \bibinfo{booktitle}{\emph{International conference on machine learning}}. PMLR, \bibinfo{pages}{6687--6696}.
\newblock


\bibitem[Wolfer and Kontorovich(2021)]%
        {wolfer2021statistical}
\bibfield{author}{\bibinfo{person}{Geoffrey Wolfer} {and} \bibinfo{person}{Aryeh Kontorovich}.} \bibinfo{year}{2021}\natexlab{}.
\newblock \showarticletitle{{Statistical estimation of ergodic Markov chain kernel over discrete state space}}.
\newblock \bibinfo{journal}{\emph{Bernoulli}} \bibinfo{volume}{27}, \bibinfo{number}{1} (\bibinfo{year}{2021}), \bibinfo{pages}{532 -- 553}.
\newblock
\href{https://doi.org/10.3150/20-BEJ1248}{doi:\nolinkurl{10.3150/20-BEJ1248}}


\bibitem[Wolsey(2021)]%
        {wolseyIntegerProgramming2021}
\bibfield{author}{\bibinfo{person}{Laurence~A. Wolsey}.} \bibinfo{year}{2021}\natexlab{}.
\newblock \bibinfo{booktitle}{\emph{Integer Programming} (\bibinfo{edition}{second edition} ed.)}.
\newblock \bibinfo{publisher}{Wiley}, \bibinfo{address}{Hoboken, NJ}.
\newblock
\showISBNx{978-1-119-60655-0 978-1-119-60652-9}
\showLCCN{T57.74}


\bibitem[Zhang and Neill(2016)]%
        {zhang2016identifying}
\bibfield{author}{\bibinfo{person}{Zhe Zhang} {and} \bibinfo{person}{Daniel~B Neill}.} \bibinfo{year}{2016}\natexlab{}.
\newblock \showarticletitle{Identifying significant predictive bias in classifiers}.
\newblock \bibinfo{journal}{\emph{arXiv preprint arXiv:1611.08292}} (\bibinfo{year}{2016}).
\newblock


\end{thebibliography}


\appendix
\section*{Appendix}
\section{Experiment details}
Each experiment was run on an internal cluster. Each node had 32 GB of RAM and either an AMD EPYC 7543 or an Intel Xeon Scalable Gold 6146 processor, based on availability.

\subsection{Datasets description}
\label{app:data}

The dataset statistics are in Table \ref{tab:datasets}. Note the Hawaii X Maine dataset for an illustration of the difficulty of assessing the subgroup bias. It is quite difficult to evaluate the bias of more than 400,000 times more groups than there are samples.
We utilize 2 types of datasets, both constructed from the US Census data - American Community Survey (ACS). The first 5 (with names starting with ACS) represent tasks constructed by the creators of the folktables library \cite{ding2021retiring}. 
%
The datasets of the other type are constructed by us using the folktables API from the same data source. We select five pairs of distinctive US states and compare their data distributions. We simply combine a randomly sampled half of the available data for each state to form our dataset. Each dataset has the same attributes listed in Table \ref{tab:attrs}. 

All used data is 1-year horizon data from the survey year 2018. We work with categorical or binary data, where we binarize continuous values into 10 equally spaced bins. We semantically reduce the cardinality of the Place of birth attribute from 221 to 6, based on continents (and keeping the US separate). 

\begin{table}
    \centering
    \caption{Datasets used in experiments. Those starting with ACS are the original tasks provided by the folktables library \cite{ding2021retiring}. (CA) means that the data is taken from the state of California. The bottom half of the datasets compares the population of US states using the US Census data. We compare states using 14 protected attributes, listed in Table \ref{tab:attrs}. 
    Datasets are in the order in which they appear in the results. 
    The number of subgroups may change even for the same set of attributes when some categorical value is missing from the data for one dataset/task and not for the other.
    }
    \resizebox{\linewidth}{!}{
    \begin{tabular}{lccc}
    \toprule
         Dataset name & \# samples & \# protected & \# subgroups \\
         \midrule
         ACSIncome (CA) & 195,665 & 4 & 1,229 \\
         ACSPublicCoverage (CA) & 138,554 & 12 & 56,008,799 \\
         ACSMobility (CA) & 80,329 & 11 & 11,567,699 \\
         ACSEmployment (CA) & 378,817 & 11 & 18,381,599 \\
         ACSTravelTime (CA) & 172,508 & 6 & 47,339 \\
         \midrule
         Hawaii X Maine & 13,837 & 14 & 5,772,124,799 \\
         California X Wyoming & 192,278 & 14 & 5,772,124,799 \\
         Mississippi X New Hampshire & 21,452 & 14 & 5,772,124,799 \\
         Maryland X Mississippi & 44,482 & 14 & 5,772,124,799 \\ 
         Louisiana X Utah & 37,595 & 14 & 5,772,124,799 \\
        \bottomrule
    \end{tabular}}
    \label{tab:datasets}
\end{table}

\subsubsection{Protected attributes}
\label{app:protected}
Below, we enumerate the attributes for each dataset. 
More details about the attributes are in Table \ref{tab:attrs}.

\begin{itemize}
    \item \textbf{ACSIncome (CA)}: \texttt{AGEP}, \texttt{POBP}, \texttt{SEX}, \texttt{RAC1P}
    \item \textbf{ACSPublicCoverage (CA)}: \texttt{AGEP}, \texttt{SEX}, \texttt{DIS}, \texttt{CIT}, \texttt{MIL}, \texttt{ANC}, \texttt{NATIVITY}, \texttt{DEAR}, \texttt{DEYE}, \texttt{DREM}, \texttt{FER}, \texttt{RAC1P}
    \item \textbf{ACSMobility (CA)}: \texttt{AGEP}, \texttt{SEX}, \texttt{DIS}, \texttt{CIT}, \texttt{MIL}, \texttt{ANC}, \texttt{NATIVITY}, \texttt{DEAR}, \texttt{DEYE}, \texttt{DREM}, \texttt{RAC1P}
    \item \textbf{ACSEmployment (CA)}: \texttt{AGEP}, \texttt{DIS}, \texttt{SEX}, \texttt{CIT}, \texttt{MIL}, \texttt{ANC}, \texttt{NATIVITY}, \texttt{DEAR}, \texttt{DEYE}, \texttt{DREM},  \texttt{RAC1P}
    \item \textbf{ACSTravelTime (CA)}: \texttt{AGEP}, \texttt{SEX}, \texttt{DIS}, \texttt{RAC1P}, \texttt{CIT}, \texttt{POVPIP}
    \item \textbf{State X State}: Every dataset with pairs of states has all 14 attributes shown in Table \ref{tab:attrs}. 
\end{itemize}

\begin{table*}
    \centering
    \caption{Protected attributes. Most attributes are commonly considered protected, with multiple being related to race and nationality. We consider Military service protected, due to existing laws against discrimination based on veteran status. The rationale behind protecting the feature of recently giving birth is linked to not uncommon discrimination in the workplace based on the prospect of women being of fertile age. For deeper insight, refer to the official documentation of the ACS data source. \vspace{0.1in}}
    \begin{tabular}{llll}
    \toprule
         ACS Code & Name & Number of values & Reduced number of values \\
         \midrule
        \texttt{SEX} & Sex & 2 & 2 \\
        \texttt{RAC1P} & Race & 9 & 9 \\
        \texttt{AGEP} & Age & continuous & 10 (binned) \\
        \texttt{POBP} & Place of birth & 221 & 6 (grouped to continents and the USA) \\
        \texttt{DIS} & Disability & 2 & 2 \\
        \texttt{CIT} & Citizenship & 5 & 5 \\
        \texttt{MIL} & Military service & 5 & 5 \\
        \texttt{ANC} & Ancestry & 4 & 4 \\
        \texttt{NATIVITY} & Foreign or US native & 2 & 2 \\
        \texttt{DEAR} & Difficulty hearing & 2 & 2 \\
        \texttt{DEYE} & Difficulty seeing & 2 & 2 \\
        \texttt{DREM} & Cognitive difficulty & 3 & 3 \\
        \texttt{FER} & Gave birth last year & 3 & 3 \\
        \texttt{POVPIP} & Income, relative to the Poverty threshold & continuous & 10 (binned) \\
        \bottomrule
    \end{tabular}
    \label{tab:attrs}
\end{table*}

\section{Further results}
\label{app:more_res}

\begin{figure*}
    \centering
    \includegraphics[width=\linewidth]{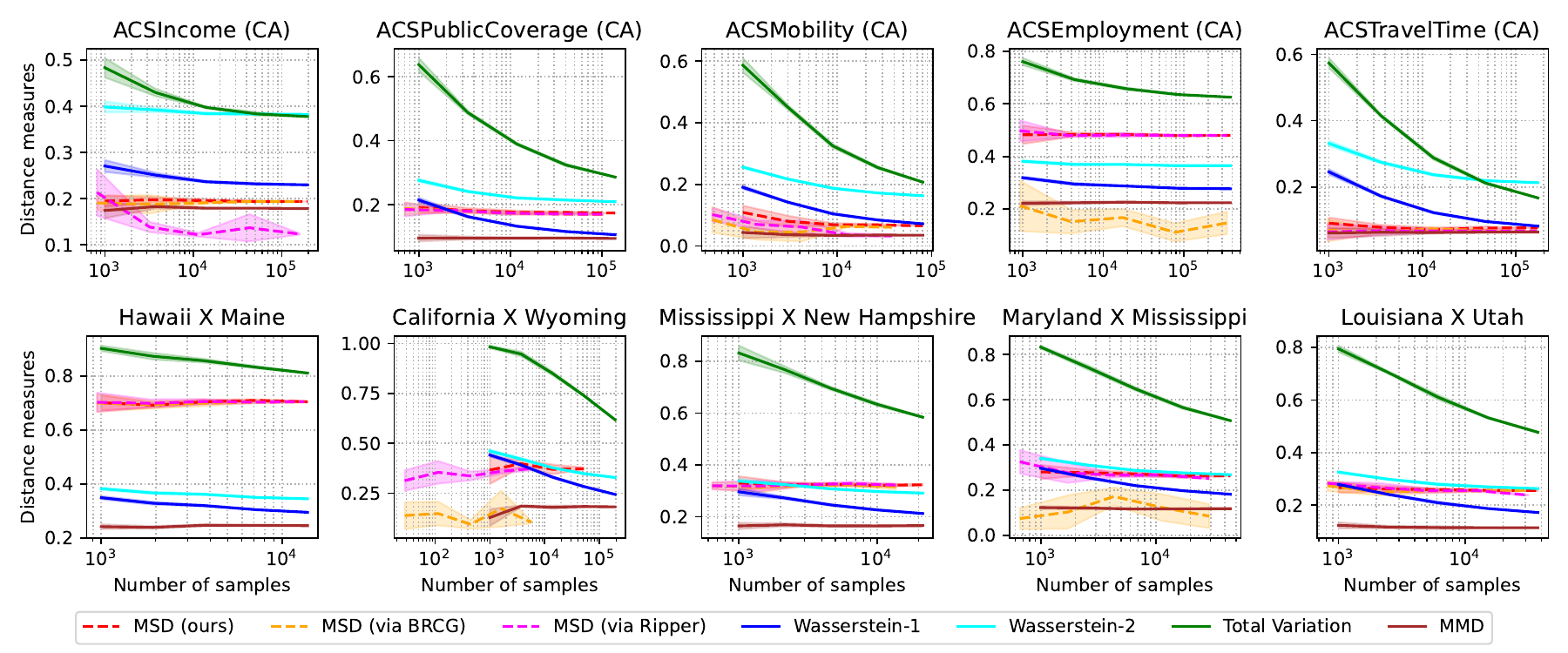}
    \caption{Original distances for all methods.}
    \Description{
    The figure contains 10 plots in 2 rows; on top are the convergence results for folktables datasets, and on the bottom are results for 5 state pair comparisons. The proposed MSD and MMD are the only ones with stable performance.
    }
    \label{fig:base}
\end{figure*}

\subsection{Original distances}
\label{app:base_distances}
In addition to the results in the main body of the paper, we present the results on the true distances, before ``normalization'' by the best-estimated distance, in Figure \ref{fig:base}. It is difficult to compare the methods due to the various distances having different interpretations. One might just comment on the distances separately. 
For example, note the range of estimates of TV; oftentimes, the value changes by around 0.3 from the first to the last estimate, which is essentially a third of the range of feasible values TV can take. 

Also, one can notice that Wasserstein metrics and the MMD are not well-suited for the evaluation of maximal intersectional bias. When comparing results on Hawaii X Maine with Mississippi X New Hampshire. Wasserstein metrics and MMD estimates differ by around 0.05 between the two datasets, while $\MSDc$ distance differs by around 0.4, which is more than half of the value for Hawaii X Maine. In other words, $\MSDc$ reports vastly different distances for the two datasets, one being twice as close, while Wasserstein metrics and MMD report the two datasets having much more similar distances, differing at most by a quarter, in the case of MMD. This is expected since $\MSDc$ finds a maximally (dis)advantaged subgroup. 

Finally, notice the difference in distance estimates between the ACSIncome and the Louisiana X Utah setups. The $\MSDc$ estimate is higher in the second case, while Wasserstein metrics, much like MMD, report higher distance in the first case. In practice, utilizing the Wasserstein or MMD could lead to choosing a classifier that seems less biased, but has a subgroup with notably higher disparity compared to all subgroup disparities of the rejected classifier.

\begin{figure*}
    \centering
    \includegraphics[width=\linewidth]{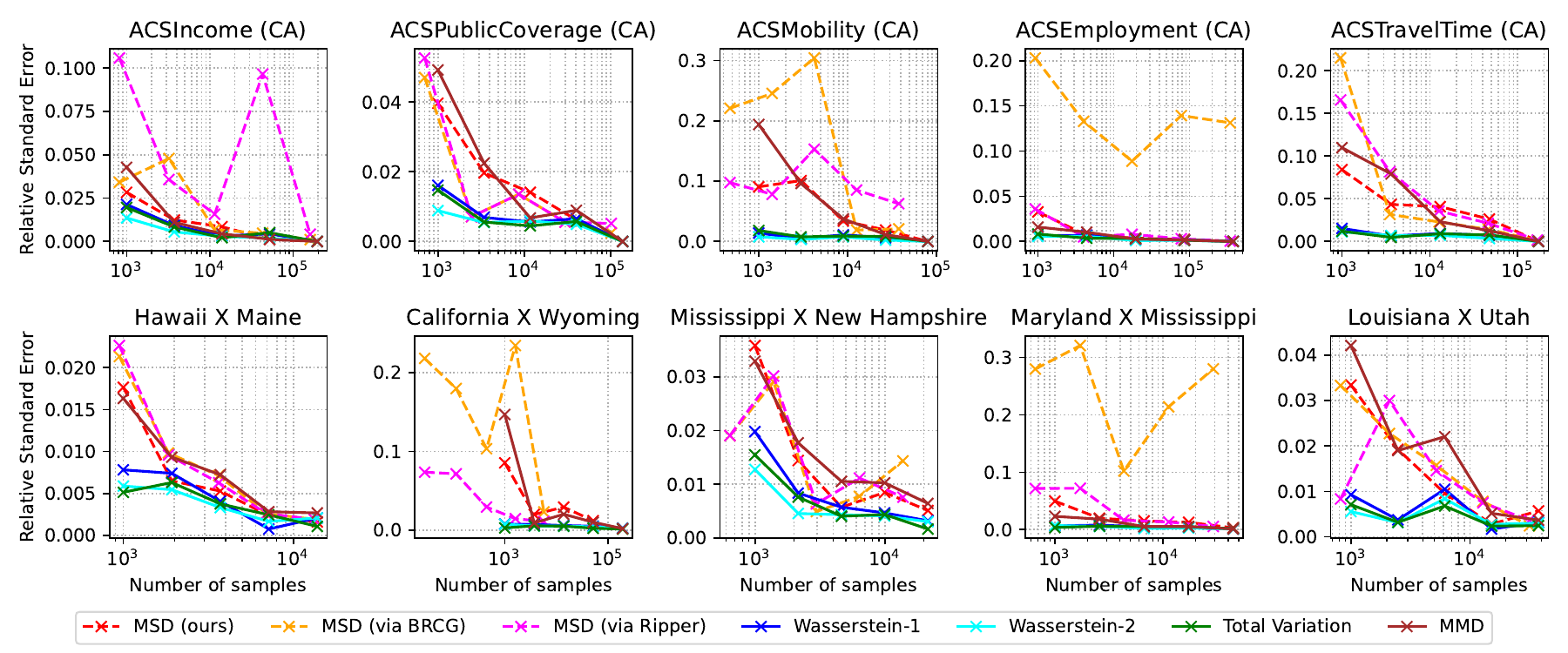}
    \caption{Relative standard error of all methods.}
    \Description{
    The figure contains 10 plots in 2 rows; on top are the relative standard errors for folktables datasets, and on the bottom are errors for 5 state pair comparisons. The Wasserstein metrics and Total Variation have the lowest errors throughout, BRCG and Ripper have errors somewhat unstable, and the MSD and MMD have similar errors, decreasing with the number of samples.
    }
    \label{fig:RSE}
\end{figure*}

\subsection{RSE}
Next, we present the results by evaluating and plotting the Relative Standard Error (RSE) in Figure \ref{fig:RSE}. It shows the stability of the estimate over various seeds. The methods evaluating $\MSDc$ seem to generally struggle more, which might be a feature of the measure, taking the supremum rather than some mean value. This could likely make the measure more ``volatile'' when using different seeds. 
However, note that MMD has comparable RSE, despite being a measure computed as a mean value.

\subsection{List of subgroups}
\label{app:subgroup_list}
To showcase the interpretability of our metric in bias detection, we list the subgroups found for each dataset. We present the globally optimal results
for the highest number of samples.
For the original datasets on the state of California, 
the subgroups are the same across all seeds:
\begin{itemize}
    \item \textbf{ACSIncome (CA)} - The youngest portion of people (\texttt{17 <=  AGEP < 25}).
    \item \textbf{ACSPublicCoverage (CA)} - People without a disability who never served in the military (\texttt{DIS = 2 AND MIL = 4}).
    \item \textbf{ACSMobility (CA)} - People who report having a single ancestry, were born in the US, do not have cognitive difficulty, difficulty seeing, and who never served in the military (\texttt{ANC = 1 AND CIT = 1 AND DEYE = 2 AND DREM = 2 AND MIL = 4}).
    \item \textbf{ACSEmployment (CA)} - People without a disability who never served in the military (\texttt{DIS = 2 AND MIL = 4}).
    \item \textbf{ACSTravelTime (CA)} - White female persons (\texttt{RAC1P = 1 AND SEX = 2}).
\end{itemize}

On datasets comparing US states, the globally optimal subgroups sometimes differ for the highest sample size. This is due to taking a random half of each state's data, which means that the full datasets can be different for different seeds, unlike the ACS data. 
See Section \ref{app:data} for more details. 
The recovered subgroups are as follows:
\begin{itemize}
    \item \textbf{Hawaii X Maine} - White people (\texttt{RAC1P = 1}).
    \item \textbf{California X Wyoming}
    \begin{itemize}
        \item White people, native to the USA (\texttt{NATIVITY = 1 AND RAC1P = 1}). 
        \item White people, born in the US, without difficulty seeing (\texttt{CIT = 1 AND DEYE = 2 AND RAC1P = 1}, for one seed). 
        \item White people, native to the USA, with place of birth in the US (\texttt{NATIVITY = 1 AND POBP = 0 AND RAC1P = 1}, for one seed).
        \item White people, born in the US, with the place of birth in the US (\texttt{CIT = 1 AND POBP = 0 AND RAC1P = 1}, for one seed).
    \end{itemize}
    \item \textbf{Mississippi X New Hampshire} 
    \begin{itemize}
        \item Black people, native to the USA (\texttt{NATIVITY = 1 AND RAC1P = 2}, for three seeds).
        \item Black people, US-born, native to the USA (\texttt{CIT = 1 AND NATIVITY = 1 AND RAC1P = 2}, for two seeds).
    \end{itemize} 
    \item \textbf{Maryland X Mississippi} - People with the highest income-to-poverty ratio (\texttt{450.94509 <= POVIP < 501.0501}).
    \item \textbf{Louisiana X Utah} 
    \begin{itemize}
        \item Black people (\texttt{RAC1P = 2}, for three seeds).
        \item Black people, native to the USA (\texttt{NATIVITY = 1 AND RAC1P = 2}, for two seeds).
    \end{itemize}
\end{itemize}





\subsection{Synthetic experiments}
We performed experiments on synthetic data with known MSD of a subgroup defined on 4 binary features out of 10 in total (distributions were otherwise similar, as in Figure \ref{fig:intersectional_fairness}). Our method achieved a mean error of 0.003 (4\%) in estimating true MSD with just $10^4$ samples. Note that this includes sampling error. Only 1 of 5 seeded runs did not recover the exact subgroup description. It did not include a single feature, returning a superset of the true subgroup.

\end{document}